\documentclass{article}

\usepackage[preprint]{corl_2023} 

\usepackage{tabularx}
\usepackage{graphicx}
\usepackage{latexsym}
\usepackage{amssymb}
\usepackage{CJKutf8}
\usepackage[export]{adjustbox}
\newcommand{\larg}{$\mbox{LARG}^2$}

\title{ {\huge \larg}, Language-based Automatic Reward and Goal Generation}

\author{
  Julien Perez\\
  Naver Labs Europe\\
  France\\
  \texttt{Julien.Perez@naverlabs.com} \\
  \And
  Denys Proux \\
  Naver Labs Europe\\
  France\\
  \texttt{Denys.Proux@naverlabs.com} \\
  \AND
  Claude Roux \\
  Naver Labs Europe\\
  France\\
  \texttt{Claude.Roux@naverlabs.com} \\
  \And
  Michael Niemaz \\
  Naver Labs Europe\\
  France\\
  \texttt{Michael.Niemaz@naverlabs.com} \\
}

\begin{document}
\maketitle


\begin{abstract}
Goal-conditioned and Multi-Task Reinforcement Learning (GCRL and MTRL) address numerous problems related to robot learning, including locomotion, navigation, and manipulation scenarios.
Recent works focusing on language-defined robotic manipulation tasks have led to the tedious production of massive human annotations to create dataset of textual descriptions associated with trajectories.
To leverage reinforcement learning with text-based task descriptions, we need to produce reward functions associated with individual tasks in a scalable manner.
In this paper, we leverage recent capabilities of Large Language Models (LLMs) and introduce \larg, Language-based Automatic Reward and Goal Generation, an approach that converts a text-based task description into its corresponding reward and goal-generation functions
We evaluate our approach for robotic manipulation and demonstrate its ability to train and execute policies in a scalable manner, without the need for handcrafted reward functions.
\end{abstract}



\section{Introduction}

In the field of robotic manipulation, decision models are evolving from optimal control approaches towards policy learning through Multi-task and Goal-Conditioned Reinforcement Learning~\citep{Huang_monolog_2022}.
Following this line of work, multi-modal task definition~\citep{Jiang_VIMA_2022, Shah_LMNAV_2022}, associated with reasoning and action planning abilities facilitated by LLMs~\citep{Huang_zeroshotplanner_2022}, enables agents to adapt to real-world uncertainty which is hardly handled with traditional robotic control.

Several strategies, such as behavioral cloning~\citep{Tai2016ASO, Kumar2022ShouldIR}, transfer learning~\citep{Stber2018FeatureBasedTL, Wiese2021TransferLF, Weng2020MultiModalTL}, or interactive learning ~\citep{Kelly2018HGDAggerII, Chisari2021CorrectMI, Faulkner2020InteractiveRL}, have been proposed for that matter. 
While promising, these former approaches are hardly scalable as they require human demonstrations or handcrafted trajectories.
In fact, \textbf{the difficulty of connecting textual descriptions of tasks with their associated computational goals and reward functions is what has led to these unscalable solutions}.

In this paper, we introduce \larg, Language-based Automatic Reward and Goal Generation.
For a given sequential decision task, our method simply automates the generation of its corresponding reward function from a textual descriptions of it.
We leverage common sense and reasoning capabilities offered by recent large language models in terms of text understanding and source code generation and embodied our approach in two settings.
First, for a given environment, our approach samples the target goal conditioned by a textual description of a task which allows to train a corresponding policy using Goal-Conditioned Reinforcement Learning.
Second, we directly generate executable reward functions according to the same task,  which allows to train a corresponding policy using Multi-Task Reinforcement Learning.
Finally, we evaluate both settings of \larg \hspace{0.05cm} over a set of language-formulated tasks in a tabletop manipulation scenario.

\section{Preliminaries}

Reinforcement Learning considers an agent which performs sequences of actions in a given environment to maximize a cumulative sum of rewards. 
Such problem is commonly framed as Markov Decision Processes (MDPs): $M = \{S, A, T, \rho_0, R\}$ ~\citep{Sutton2005ReinforcementLA, Mnih2016AsynchronousMF, Lillicrap2016ContinuousCW}. 
The agent and its environment, as well as their interaction dynamics, are defined by the first components $\{S, A, T, \rho_0\}$, where $s \in S$ describes the current state of the agent-environment interaction and $\rho_0$ is the distribution over initial states.
The agent interacts with the environment through actions $a \in A$. 
The transition function $T$ models the distribution of the next state $s_{t+1}$ conditioned with the current state and action $T: p(s_{t+1}| s_t, a_t)$.
Then, the objective of the agent is defined by the remaining component of the MDP, $R: S \rightarrow \mathbb{R}$.
Solving a Markov decision process consists in finding a policy $\pi: S \rightarrow A$ that maximizes the cumulative sum of discounted rewards accumulated through experiences.

In the context of robotic manipulation, a task commonly consists in altering the environment into a targeted state through selective contact ~\citep{rlrobotmanip_Gu_2017}.
So, tasks are expressed as $g = (c_g, R_G)$ pair where $c_g$ is a goal configuration such as Cartesian coordinates of each element composing the environment or a textual description of it, and $R_G: S \times G \rightarrow \mathbb{R}$ is a goal-achievement function that measures progress towards goal achievement and is shared across goals.
A goal-conditioned MDP is defined as : $M_g = \{S, A, T , \rho_0, c_g, R_G\}$ with a reward function shared across goals. 
In multi-task reinforcement learning settings, an agent solves a possibly large set of tasks jointly.
It is trained on a set of rewards associated with each task. 
Finally, goals are defined as constraints on one or several consecutive states that the agent seeks to satisfy ~\citep{Plappert2018MultiGoalRL, Nair2018VisualRL, OpenAI2021AsymmetricSF}.

\section{Related work}


\subsection{Challenges of reward definition and shaping}

A sequential decision task requires defining an informative reward function to enable the reinforcement learning paradigm.
Reward shaping consists in designing a function in an iterative process incorporating elements from domain knowledge to guide policy search algorithms.
Formally, this can be defined as $R' = R + F$, where $F$ is the shaping reward function, and $R'$ is the modified reward function \cite{dorigo1994robot,randlov1998learning}. 
As a main limitation, a reward function needs to be manually designed for each task. 
For instance~\citep{Brohan2022RT1RT} leveraged large number of human demonstrations and specific handcrafted definitions of tasks to train a robotic transformer. 
However, as MTRL can deal with a large set of goals and tasks to implement, such an approach becomes hardly scalable.
In this work, we study how to leverage LLMs and the common-sense and prior knowledge embedded in such models to automate the paraphrasing of tasks and the generation of associated reward functions.

\subsection{Large Language models for control}

Goal-conditioned reinforcement learning has recently been successfully considered in the domain of robot control with textual descriptions of tasks.
\citep{Shah_LMNAV_2022} has combined a text encoder, a visual encoder, and visual navigation models, to provide text-based instructions to a navigating agent. 
This paradigm has been further developed in~\citep{Huang_monolog_2022} using the capabilities of large language models to support action planning, reasoning, and internal dialogue among the various models involved in manipulation tasks. 
Along these lines, ~\citep{Liang_codeaspolicy_2022} proposes to use the code generation capabilities of sequence-to-sequence language models to transform user instructions into a code-based policy.
However, it involves an interactive design process for a hard-coded policy, rather than a task-conditional learning process and it does not benefit from a policy trained through reinforcement learning.
Finally, ~\citep{kwon_rewarddesign_2023} proposes to use language to control the reward value in a feedback loop to adapt robot behavior. 
In this case, no code is generated and only the reward signal is modified according to user guidelines. 
Furthermore, it requires constant feedback from the user which makes it hardly scalable.

The closest to our work is~\citep{Colas2020LanguageConditionedGG} which proposes to automatically derive goals from a textual description of the task. 
However, the language remains limited to the logical descriptions of the expected configuration of the scene and the goal is reduced to a finite set of eligible targets. 
Our approach allows using natural language beyond logical forms, grounded with reasoning capabilities and enriched with common-sense captured in large pre-trained language models. 
Also related, ~\citep{2020-Sigaud-goalgeneration} propose to train a conditional variational auto-encoder to create a language-conditioned goal generator. 
However, it assumes the existence of pre-trained goal-conditioned policies. 
Furthermore, no LLM background knowledge and code generation is considered to achieve this objective.

As far as our knowledge goes, our work is the very first attempt to leverage in-context prompting to enable training goal-condition and multi-task reinforcement learning agents with a textual description of the task in a scalable manner so that goal sampling and reward codes can be adapted according to the user-specific objectives.

\section{Language-based Automatic Reward and Goal Generation}

\begin{figure*}[t!]
    \centering
    \includegraphics[width=\linewidth]{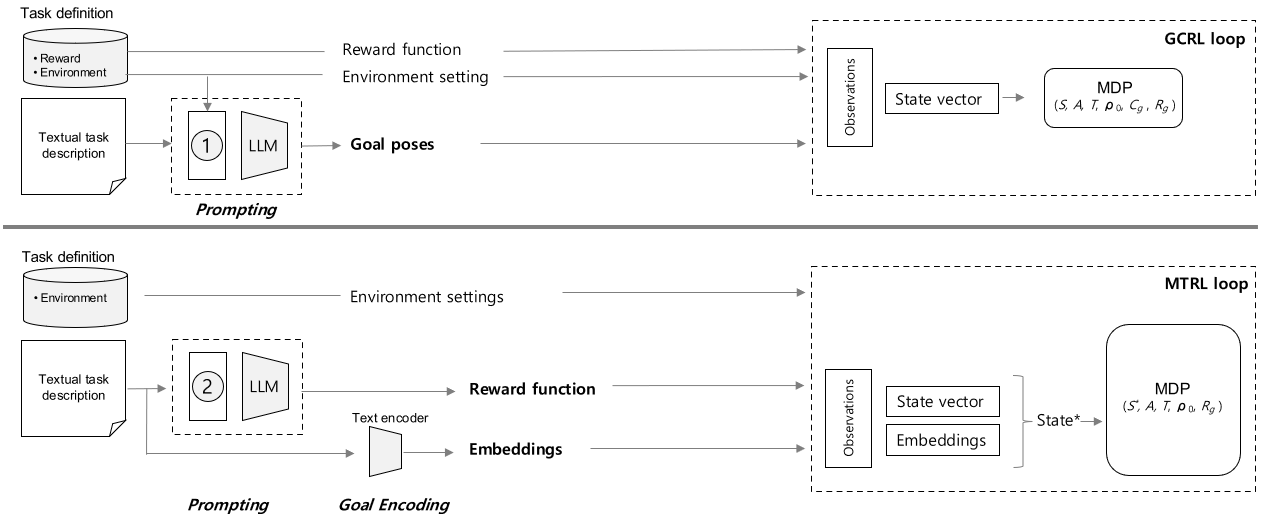}
    \caption{\larg transforms a textual task description into either 1) a goal to be used as input of a pre-existing reward function for GCRL, or 2) into a reward function for MTRL. 
    We leverage pre-trained LLM with dedicated prompts for our generation procedure. For GCRL, the goal is appended to the state description given as input to the policy. For MTRL the text-based task description is encoded using a pre-trained language model to complement the state vector.} 
    \label{fig:1}
\end{figure*}

We introduce two methods using LLMs for handling textual descriptions of tasks as illustrated in Figure \ref{fig:1}.
First, \textit{Automatic Goal Generation} uses LLMs to produce executable code that generates goals ($c_g$) to be used as parameters of a predefined goal-conditioned reward function ($R_G$).
In this case, the trained policy can take as input a goal generated from such an LLM-produced function.
Second, \textit{Automatic Reward Generation} uses LLMs to generate a reward function associated with a given textual description of a task.
In this second approach, the policy takes as input the textual description of the task.
After this generation phase, we leverage the off-the-shelf multi-task reinforcement learning framework to train agents.
The following sections describe our two approaches and detail their corresponding prompting strategies.

\subsection{Automatic Goal Generation}
\label{goalgeneration}

Our first objective is to translate a textual task description with its constraints and guidelines into a goal.
As a pre-requisite, we assume the existence of categories of tasks along with their environment settings and associated reward functions.
These reward functions are parameterized with a goal.
In tabletop robotic manipulation scenarios, a task consists in re-arranging a set of objects composing the scene. 
We assume that the goal is the set of target poses for all objects. 
Then, the reward function incorporates environment-dependent reward terms and Euclidian distance between the current pose of the objects and the target pose.
Goals generated by \larg \hspace{0.05cm} are used in a GCRL learning setting to compute the reward signal at each step.
The prompt design $p$, described in section ~\ref{Prompt}, generates a function $f$ returning eligible values for the targeted task such as $f \rightarrow c_g$ where $c_g = [goal\_ values]$.

\subsection{Automatic Reward Generation}
\label{rewardgeneration}

The second utilization of \larg \hspace{0.05cm} generates the executable source code of a reward function according to a textual description of a task.  

While Large Language Models could possibly support the full generation of complex reward functions, we propose to identify different parts in such a function, some being task-independent and others closely related to the task definition.
In robotic manipulation, common task-independent components address bonuses for lifting the objects or penalties for the number of actions to achieve a given purpose.
Task-dependent components are driven by the textual task description and align constraints with penalties and guidelines with bonuses.
Both components are combined in a global reward function.
Considering current LLM limitations, our experiments highlighted the fact that it is more efficient to perform such a decomposition to decrease the complexity of the generated code.    

For the composition of the global reward function, we consider the existence of predefined categories of tasks with their environments, formalized using languages such as YAML~\footnote{https://yaml.org/} or Python~\footnote{https://www.python.org/} providing task dependent reward components such as what exists in repositories like Isaac\_Gym~\footnote{https://developer.nvidia.com/isaac-gym}. 
\larg \hspace{0.05cm} aligns specific textual task description with a related task category and generates the task-dependent part of the reward function to form the global function.

\subsection{Task-encoding and Policy}

For GCRL, the input of the policy is composed of the environment state and the goal generated by \larg.
For MTRL, the goal of each task is replaced by a textual description of the task. 
To do this, we use a pre-trained text encoder such as Google T5~\citep{raffel2020T5, Raffel2020ExploringTL} to tokenize and encode the text into an embedding vector.
This vector is added to the state vector in the training phase to caption tasks.
This approach allows using textual descriptions of tasks as input to neural policies such as what is proposed by~\citep{Jiang_VIMA_2022}.

\section{In-context Code Generation}
\label{CodeGeneration}

\subsection{Prompting}
\label{Prompt}

We describe how the prompt is designed as a composition of textual instructions, code samples, and Docstring \footnote{https://peps.python.org/pep-0257/}.
We use the context of robotic manipulation to illustrate our approach.
While experimenting with our approach, we notice that prompt is a crucial step in our approach.

\begin{figure}[t]
   \begin{minipage}{0.6\textwidth}
     \centering
    \includegraphics[width=0.5\linewidth]{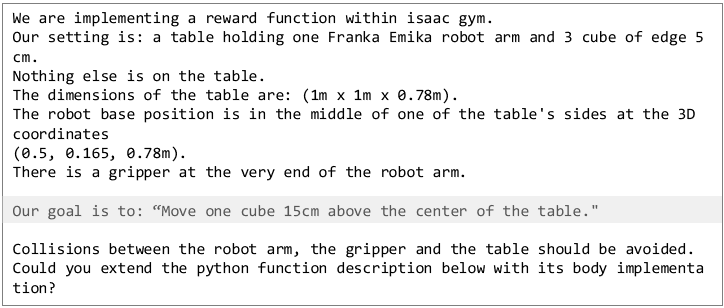}
    \includegraphics[width=0.5\linewidth]{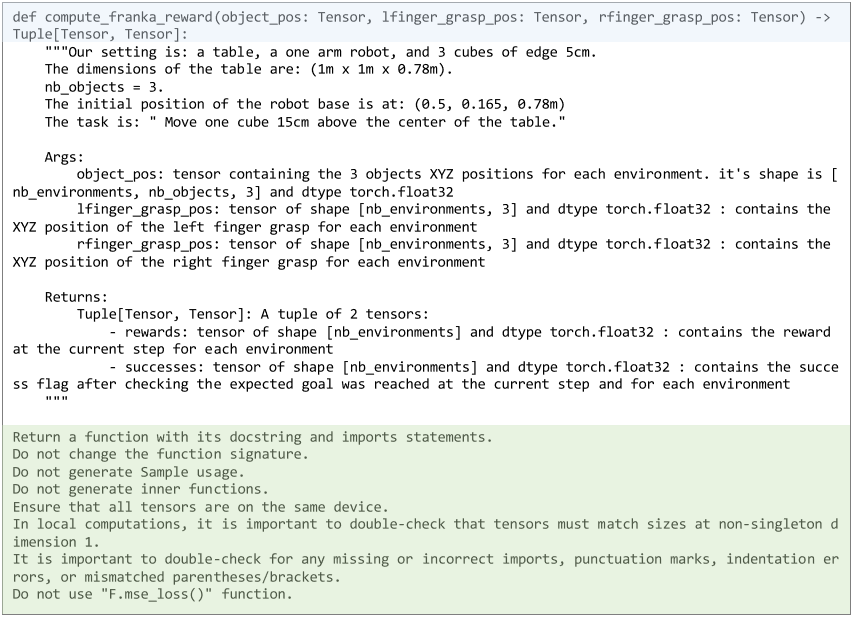}
   \end{minipage}\hfill
   \begin{minipage}{0.4\textwidth}
     \centering
     \includegraphics[width=.9\linewidth]{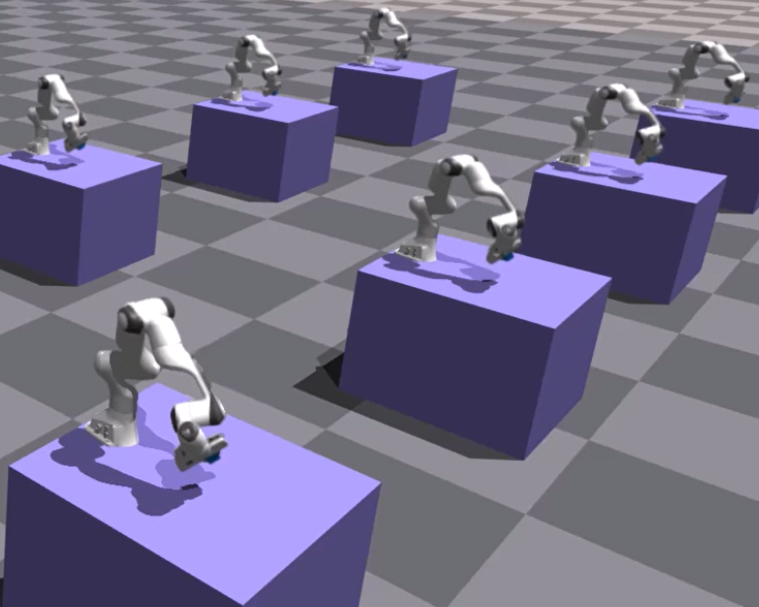}
   \end{minipage}
    \caption{Example of prompt generating a reward function and its application in our simulation. The section defining the purpose of the task is highlighted in grey. The signature of the expected function is in blue. The part in green relates to additional guidelines.}
    \label{fig:4}

\end{figure}

For both goal and reward generation, the prompt structure, as illustrated in Figure ~\ref{fig:4}, consists in four parts, organized as follows: (1) the environment description, (2) the task description, (3) the specifications of the expected function and (4) additional guidelines.
First, the environment description starts with a reference to a specific category of task and then provides additional details to be considered to accomplish the task. 
It completes the context used by the LLM with elements such as the initial environment state, robot specifics, and any other relevant information like a scene description. 
It also includes technical guidelines such as API or methods to be used.
Second, the task description is a text detailing the purpose of the task. 
It is also used to caption the task in the MTRL approach.
Third, to enforce the coherence of the generated output, we provide the signature of the expected function, either for goal or reward generation, listing all parameters along with the expected returned elements. 
This signature is completed by a Docstring detailing the role of each parameter.
Finally, we can add additional guidelines to drive the code generation. 
This addition allows for mitigating frequently observed errors due to limitations in current LLMs. 
For instance, one constraint is to push data into specific devices, e.g. in the CPU or GPU. 
Such limitations can be automatically captured by the method discussed in section ~\ref{Validation}.

To generate a large set of tasks in a scalable manner, we preserve parts 1, 3 and 4 of the prompt and use the LLM to automatically rephrase a task definition into semantically similar variants using paraphrasing.

\subsection{Code Validation and Auto-correction}
\label{Validation}

\begin{figure*}[ht!]
    \centering
    \includegraphics[width=0.8\linewidth]{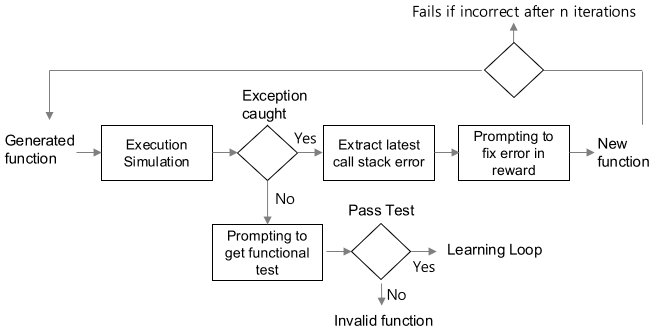}
    \caption{Diagram of the code correction loop. It leverages exceptions raised during execution to request modifications. A functional test is also requested before moving to the learning loop.}
    \label{fig:8}
\end{figure*}

The generated code isn't guarantee to meet expectations in terms of code validity or outcomes.
In such a case, further prompt iterations are performed, emphasizing the elements that need to be modified until the result converges toward expectations.
These errors mainly come from under-specified elements in the original prompt or from LLM limitations such as hallucinations~\citep{Ziwei_hallucination_2022}.
Therefore, we finalize the code generation with an automatic validation step described in Figure~\ref{fig:8} which exploits the output of the Python interpreter.

\begin{figure}[!htb]
   \begin{minipage}{0.48\textwidth}
     \centering
     \includegraphics[width=1.\linewidth]{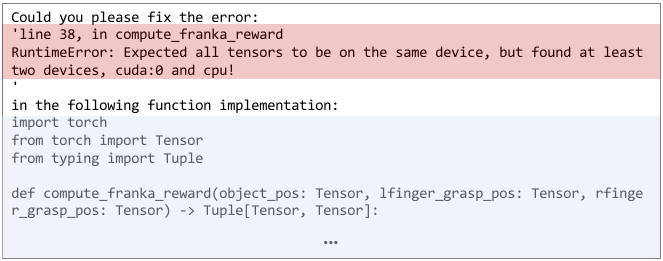}
     \caption{Prompt for the automatic code correction step. The part in blue is the code of the incorrect function generated in the previous steps and in red the output of the code interpreter.}\label{fig:9}
   \end{minipage}\hfill
   \begin{minipage}{0.48\textwidth}
     \centering
     \includegraphics[width=1.\linewidth]{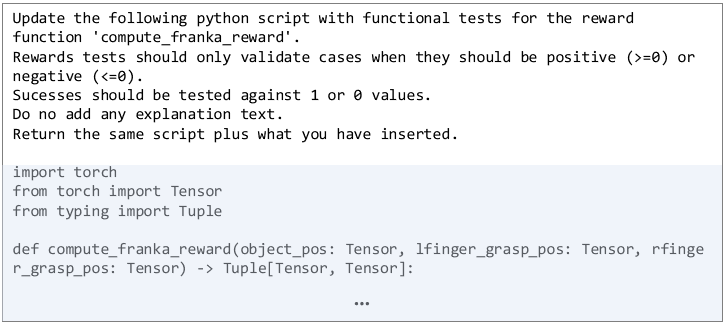}
     \caption{Prompt requesting the generation of a functional test for the  function provided, in blue, as contextual information.}\label{fig:10}
   \end{minipage}
\end{figure}

We execute the generated code on placeholder input variables and catch the exceptions raised by the Python interpreter when the code fails either to pass the syntax evaluation step or the execution step. 
We filter the thread of exceptions to only keep the latest stack and use the error message to fill a prompt requesting code modifications.
Our prompt, as illustrated in Figure ~\ref{fig:9}, contains 1) a header that requests the LLM to fix the raised exception, 2) the text of the raised exception, and 3) the code of the incorrect function.
Several iterations can be done until the code converges toward a version that can be properly executed.

Finally, once a generated function satisfy the code correction step, we use another prompt to request the LLM to generate a functional test to evaluate this first function as detailed in the appendix. 
This last validation step is intended to further filter out potentially incorrect code prior to running the training loop. 
This prompt, illustrated in Figure ~\ref{fig:10}, is composed of 1) a header requesting the LLM to generate a functional test, 2) a list of guidelines to condition the test, and 3) the code of the generated function.

\section{Experiments}
\label{experiments}

The purpose of our experiments is three-fold.
First, we evaluate \larg capabilities to generate valid goal positions for GCRL settings.
Second, we evaluate the capability of our approach to produce reward functions to train multi-task policies with language-based task description as input. 
Finally, for both cases mentioned above, we evaluate limitations of our method. 

We use a Pick and Place task family defined in the Isaac\textunderscore gym repository with a Franka Emika Panda robot arm~\footnote{https://www.franka.de/} and leverage the gpt-3.5-turbo language model~\footnote{https://platform.openai.com/docs/models/gpt-3-5} from OpenAI.
Several alternative LLMs have been experimented including StarCoder ~\footnote{https://huggingface.co/bigcode/starcoder} which has also been evaluated. 
Details of these experiments are further discussed in the appendix.
In the GCRL experiment, we evaluate goal generation on a series of 32 tasks including 27 tasks involving a single object, and 5 tasks involving 3 objects. 
A list of these tasks is provided in the appendix. 
In the MTRL experiment, we address the generation of reward functions for 9 manipulation tasks detailed in Table~\ref{tab:1cube.MTRL.listtask}. 

Regarding real-world validation concerns, both our GCRL and MTRL evaluations are conducted in a simulated environment. 
We focus on various types of rearrangement manipulation tasks and assume the capability for sim2real transfer using techniques like domain randomization. 
We chose to leverage simulated experiments to enhance the task diversity and assess the scalability of our approach, which we consider to be our primary contribution.

\subsection{Automatic goal generation}

\begin{figure}[!htb]
   \begin{minipage}{0.48\textwidth}
     \centering
     \includegraphics[width=.9\linewidth]{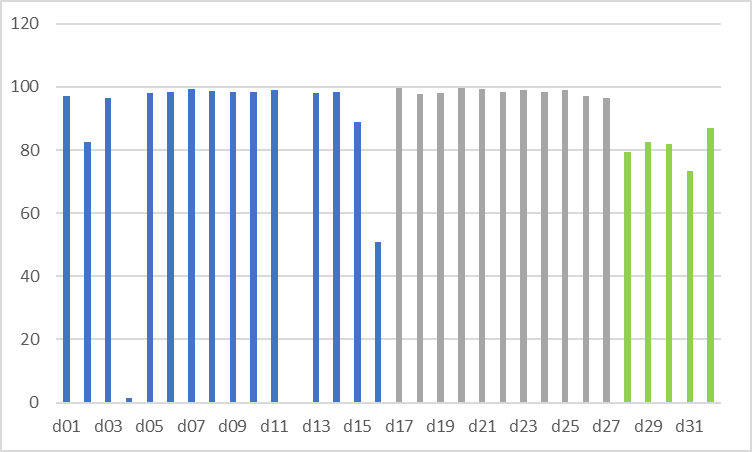}
     \caption{Success rate for GCRL manipulation tasks. Blue reflects 1 object manipulation for absolute pose whereas grey reflects relative object pose. Green relates to 3 object manipulation tasks.}
     \label{fig:70}
   \end{minipage}\hfill
   \begin{minipage}{0.48\textwidth}
     \centering
     \includegraphics[width=.8\linewidth]{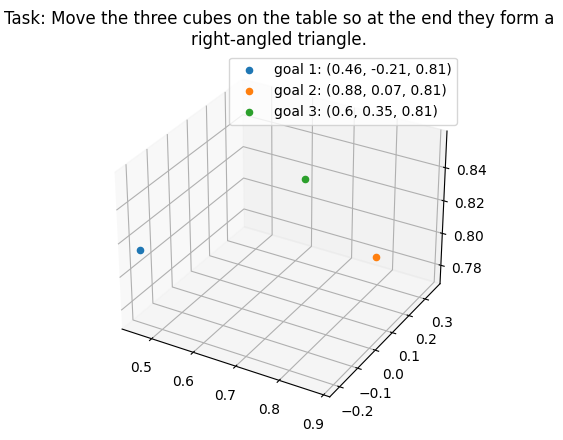}
     \caption{5 instances of the goal generation function for \textit{Task d30} which  forms a right-angled triangle.}
     \label{fig:72}
   \end{minipage}
\end{figure}

In the GCRL experiment, we use a neural policy trained beforehand using Proximal Policy Optimization ~\citep{Schulman2017ProximalPO}.
The policy takes as input the position and velocity of each joint of the robot and the respective pose of the objects composing the scene.
The policy trigger joint displacement in a $\mathbb{R}^7$ action space.
The goal information, generated by \larg \hspace{0.05cm} is used as additional input to the policy.
Figure~\ref{fig:70} provides the respective success rates for 32 manipulation tasks.

Most failures are related to the position of the goal being beyond the reachable workspace of the robotic arm. 
Nevertheless, we note numerous reasoning capabilities from the large language model to generate a code addressing textual guidelines that require common sense.
For instance in Figure ~\ref{fig:72}, goal positions need to comply with relative constraints such as: \textit{Move the three cubes on the table so at the end they form a right-angled triangle}.
From this experiment, \larg \hspace{0.05cm} demonstrated the capability to automatically produce goal positions aligned with textual task descriptions. 
Besides limitations of current LLM, 87.5\% of generated functions pass the validation test and deliver proper input to the GCRL policy to make it actionable by the robot arm.

\subsection{Automatic reward generation}

For the MTRL experiment, we encode textual descriptions of each task using a pre-trained Google T5-small language model. For each task, we use the [CLS] token embedding computed by the encoder layer of the model which is defined in $\mathbb{R}^{512}$. 
We concatenate this embedding with the state information of our manipulation environment defined in $\mathbb{R}^{7}$ and feed it into a fully connected network stack used as policy.
This policy is composed of $3$ layers using respectively, $\{512, 128, 64\}$ hidden dimensions.
Alternately, as suggested by ~\citep{Jiang_VIMA_2022}, we tested feeding the token embedding into each layer of the stack instead of concatenating it as input, but, we did not observe improvements.

For the generation of reward functions, as described in section ~\ref{rewardgeneration}, we first define the task-independent reward component for pick and place manipulation which handles bonuses for lifting the object and penalties for the number of actions to reach the objective. 
This component, which is therefore common to each task, is preset and not generated. It is added to the reward part generated by \larg \hspace{0.05cm} for each task to train the MTRL policy. Details about this process are further discussed in the appendix.

\begin{figure}[!htb]
   \begin{minipage}{0.48\textwidth}
     \centering
     \includegraphics[width=.85 \linewidth]{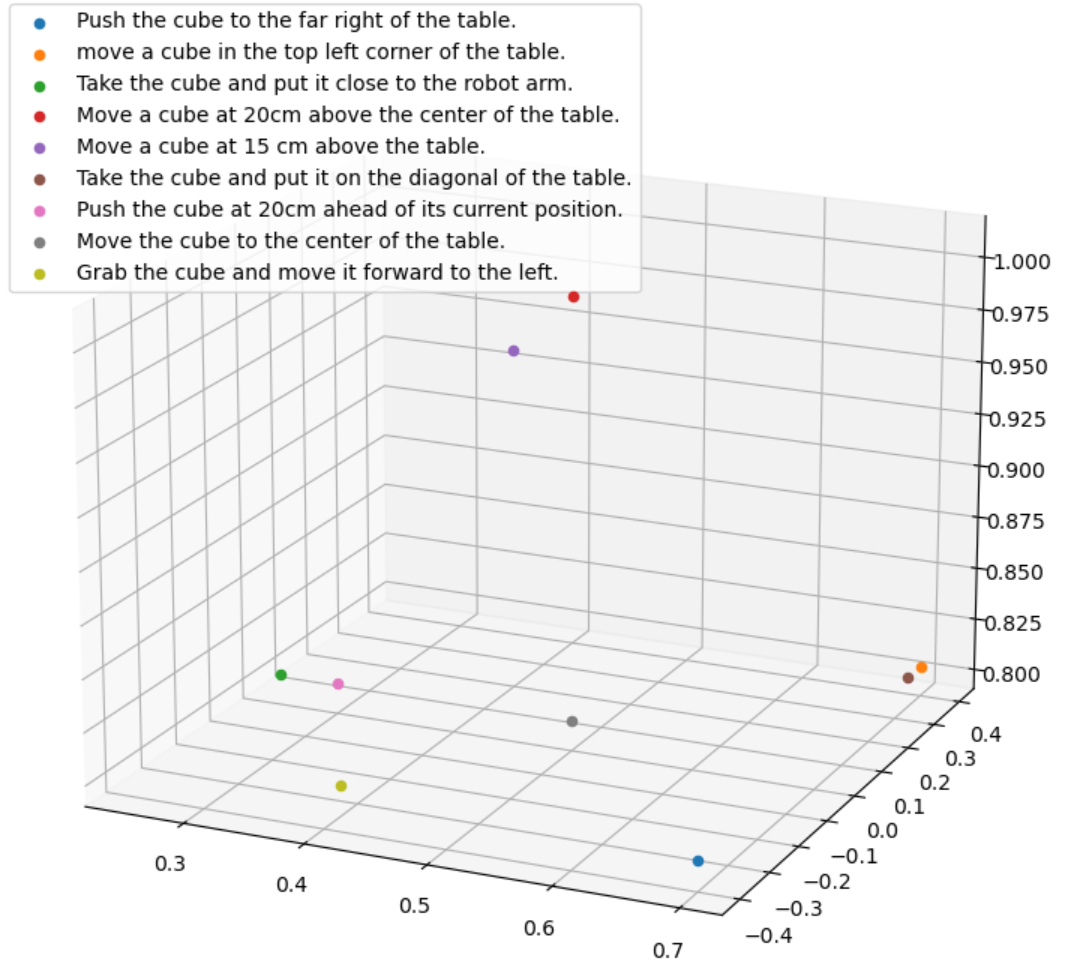}
     \caption{Generated goal position for 9 manipulation tasks.}
     \label{fig:80}
   \end{minipage}\hfill
   \begin{minipage}{0.48\textwidth}
     \centering
     \includegraphics[width=1.\linewidth]{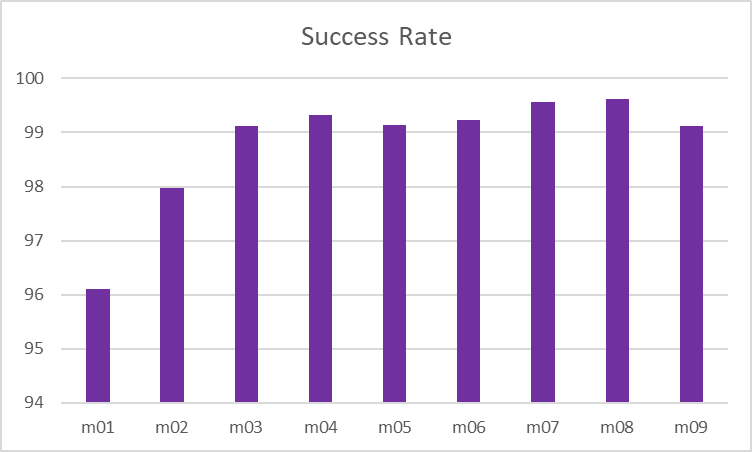}
     \caption{Success rate evaluations of MTRL over automatic reward generation.}
     \label{fig:82}
   \end{minipage}
\end{figure}

Reward functions apply goal poses generated according to the task to compute related scores. 
Table~\ref{tab:1cube.MTRL.listtask} shows the adherence of these poses with respect to task definitions using
functional tests discussed in section ~\ref{Validation} to assess this adherence.
Independently, we evaluate in Figure~\ref{fig:82} the success rate of trained policies to achieve our tasks with respect to generated goal positions.

\begin{table*}[ht!]
\tiny
\centering
\begin{tabular}{|l|p{8cm}|c|}
\hline

ID & Task & Generated Pose validity   \\
\hline
m01 & Push the cube to the far right of the table. & \checkmark  \\
m02 & Move a cube to the top left corner of the table. & \checkmark  \\
m03 & Take the cube and put it close to the robot arm. & \checkmark  \\
m04 & Move a cube at 20cm above the center of the table. & -  \\
m05 & Move a cube at 15 cm above the table. & \checkmark  \\
m06 & Take the cube and put it on the diagonal of the table. & -  \\
m07 & Push the cube at 20cm ahead of its current position. & -  \\
m08 & Move the cube to the center of the table. & \checkmark  \\
m09 & Grab the cube and move it forward to the left.  & \checkmark  \\
\hline
\end{tabular}
\caption{Manipulation tasks evaluated with \larg in MTRL settings. We assess the alignment of generated goal positions with respect to task definitions. }
\label{tab:1cube.MTRL.listtask}
\end{table*}

As a summary, \larg \hspace{0.05cm} demonstrates the capability to produce valid reward functions to successfully train MTRL policies. 
In addition, LLM can re-phrase or translate textual task definitions. 
It allows generating in a scalable manner large collections of tasks for training MTRL models with multiple paraphrases associated with each task, increasing the robustness of the approach.
\section{Limitations and future works}

Our experiments highlighted limitations of LLMs capability and reliability in converting user instructions into executable and valid code. 
It forces users to focus code generation toward specific components rather than toward broader modules. 
Even though our experiments involved highly structured information such as function signature and docstring, which somehow limits the effect of hallucination, the risk of semantic errors cannot be entirely ruled out. 
To address these limitations, the auto-correction loop described in our paper seems an effective option to be further investigated.
Nevertheless, opportunities offered by current LLMs should encourage the robotic learning community to develop the research in this domain proposed by \larg which we believe promising.
Finally, our evaluations of GCRL and MTRL are conducted in simulations, where it is possible to leverage techniques like domain randomization to demonstrate the potential for sim2real transfer. 
Simulated experiments allow us to increase task variety and  evaluate the scalability of our approach, which is our main contribution and plan to implement real-world scenario in the near future.

\section{Conclusion}

In this paper, we introduce \larg, a novel approach for task-conditioned reinforcement learning from textual descriptions.
Our method leverages the in-context learning and code-generation capabilities of large language models to complete or fully generate goal-sampling and reward functions from textual descriptions of tasks. 
For this purpose, we propose a method for automatic code validation and functional testing.
We evaluate the capability of our method to translate 32 text-based task descriptions into actionable objectives for GCRL and to 
generate rewards functions to train MTRL policies for a 7DoFs robotic arm.
Our experiment confirms the benefit of large language models for aligning textual task descriptions with generated goal and reward functions, opening the door, thanks to paraphrasing, to a scalable approach for training  Multi-Task Reinforcement Learning models.
We believe \larg opens a novel and scalable direction for training and controlling robots using textual instructions.


\clearpage


\bibliography{LARG2arxiv}



\appendix

\section{Appendix}
\label{sec:appendix}

In this section, we go through additional details about \larg \hspace{0.05cm} and discuss experiments performed to evaluate its performance both for GCRL and MTRL settings. We also provide some examples of prompts used in our experiments.

\subsection{Method}

\subsubsection{Prerequisites}

\larg \hspace{0.05cm} aims at providing a scalable method to align language-based description of tasks with goal and reward functions to address GCRL or MTRL.
It relies on code generation capabilities offered by recent Large language models. 
These LLM already capture prior background knowledge and common sens. 
In terms of coding capabilities, they leverage existing code available in repositories like GihHub \footnote{https://github.com/}.
Fully capable LLM could in theory generate proper code from pure textual descriptions.  
However, our experiments demonstrate that current LLM still benefit from additional guidelines provided as context. 
Such guidelines relate to scene understanding and function signature.
One source of information for guidelines is environment descriptions in  code repositories.
Additionally, scene understanding can be provided by exteroception components that translate images into object captions and geometric coordinates.

As a first iteration we assume the existence of a portfolio of categories of manipulation tasks defined in repositories like Isaac Gym \footnote{https://github.com/NVIDIA-Omniverse/IsaacGymEnvs}  with descriptions of  environments formalized using languages like YAML ~\footnote{https://yaml.org/} or Python ~\footnote{https://www.python.org/}. 
Accordingly, we assume that such environments also provide signatures of expected functions commented with a formalism like Docstring \footnote{https://peps.python.org/pep-0257/}.

In such a case, \larg \hspace{0.05cm} aligns a text based task description with the appropriate category of tasks and leverage  environment descriptions to build an ad-hoc prompt to be used with LLMs. 
Therefore, code generated by \larg can be seamlessly integrated into these repositories to execute the desired settings.

Textual descriptions of tasks allows to overload generic definitions of tasks available in code repositories. 
Scalability can therefore be  achieved thanks to paraphrasing. Indeed, LLMs can generate 
task definition variants on a basis of textual seeds to produce large training datasets.

\subsubsection{Generation of goal poses for GCRL}

A first application of \larg generates goals to be used as parameters of a predefined goal-conditioned reward function. 

As an example, in tabletop robotic manipulation scenarios, a pick and place task consists in re-arranging a set of objects composing a scene. 
In such a case, the goal is the set of target poses for all objects and the reward function basically compute the Euclidian distance between a current object pose and the target pose.
\larg \hspace{0.05cm} generates functions producing a set of eligible goal positions from textual task descriptions.

The prompt design used to generate the goal function is composed of the following elements: 1) the environment description, 2) the task description,  3) the specifications of the expected function and 4) optional guidelines.

\begin{figure}
    \centering
    \includegraphics[width=\linewidth]{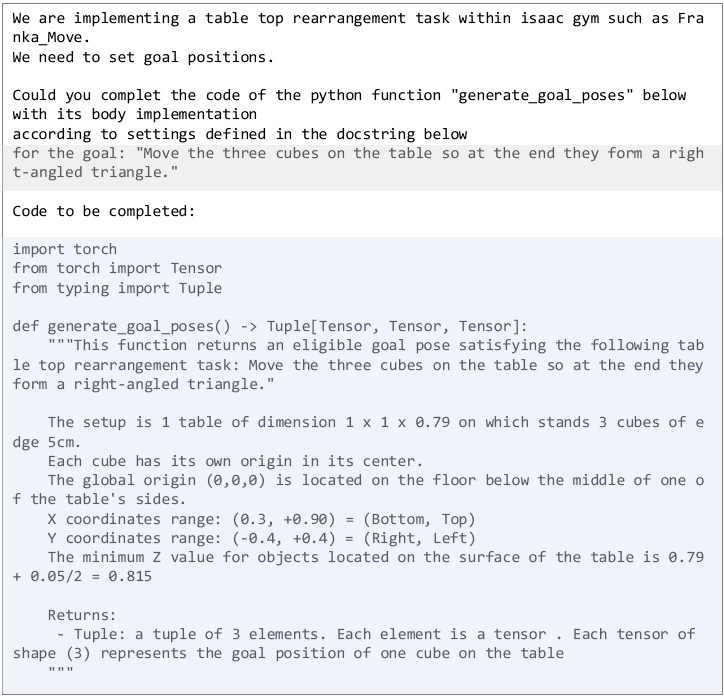}
    \caption{GCRL: Prompt requesting the generation of the goal function. The function signature is highlighted in blue and the the text-based goal description in grey. }
    \label{fig:40}
\end{figure}

\begin{figure}
    \centering
    \includegraphics[width=\linewidth]{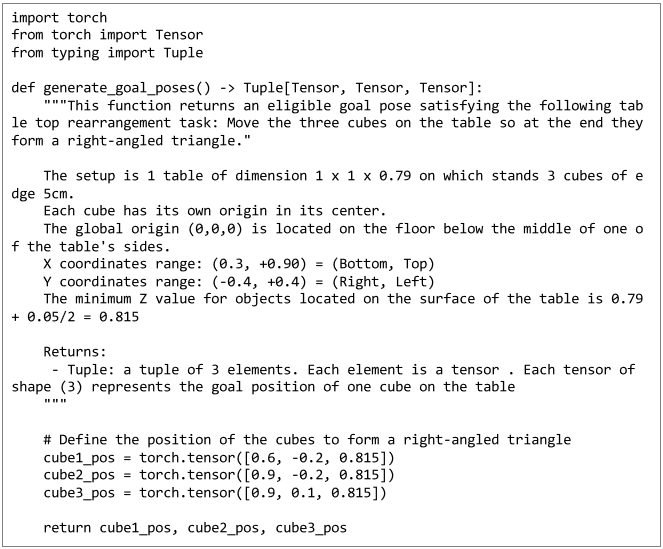}
    \caption{GCRL: Generated code for the goal pose function. }
    \label{fig:402}
\end{figure}

Figure~\ref{fig:40} illustrates our prompt design and figure~\ref{fig:402} shows the  generated code.

\newpage

\subsubsection{Generation of reward function for MTRL}
\label{app.reward}

The second utilization of \larg \hspace{0.05cm} generates the executable source code of a reward function according to a task description . 


For MTRL the policy takes as input the textual description of the task in addition to the state. 
In such as case,  goals are removed from the environment. 
However this information is required by the reward function to compute a gain.
Therefore this information is also generated by \larg \hspace{0.05cm} according to the provided task description.

For the reward function itself, we separate components which are task independent from those which are task dependent.
In robotic manipulation, task agnostic components address bonuses for lifting the objects or penalties for the number of actions to reach the goal.
Due to known limitations in current LLM, we focus \larg \hspace{0.05cm} on generating the part of the reward that depends on the specific guidelines and constraints defined in textual definitions.

The prompt structure used for generating the reward function is similar to the one used for goal generation. It is  composed of   1) the environment description, 2) the task description,  3) the specifications of the expected function and 4) optional guidelines.
However, in this case the function specification contains the signature of the expected reward function.

\begin{figure}[h]
    \centering
    \includegraphics[width=\linewidth]{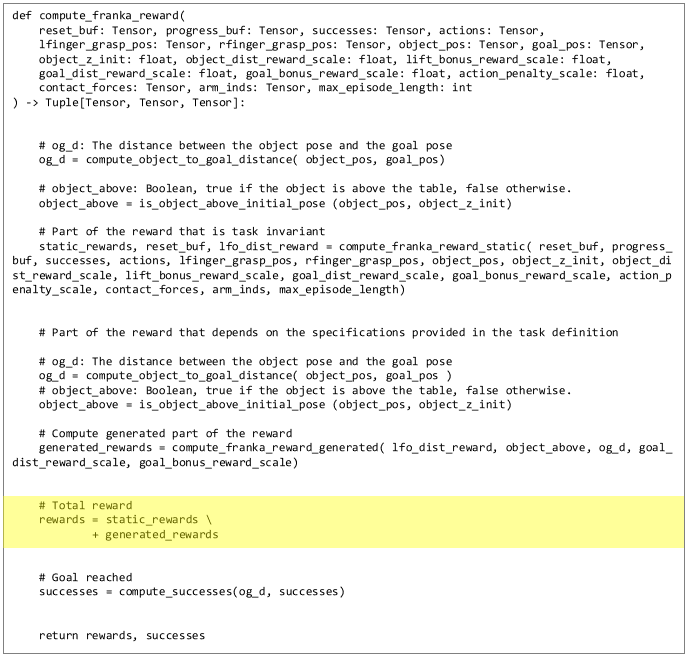}
    \caption{MTRL: Code of a global reward function combining a task independent and task dependent component (highlighted section in yellow). }
    \label{fig:app.MTRL.84.global}
\end{figure}

\begin{figure}[ht!]
    \centering
    \includegraphics[width=\linewidth]{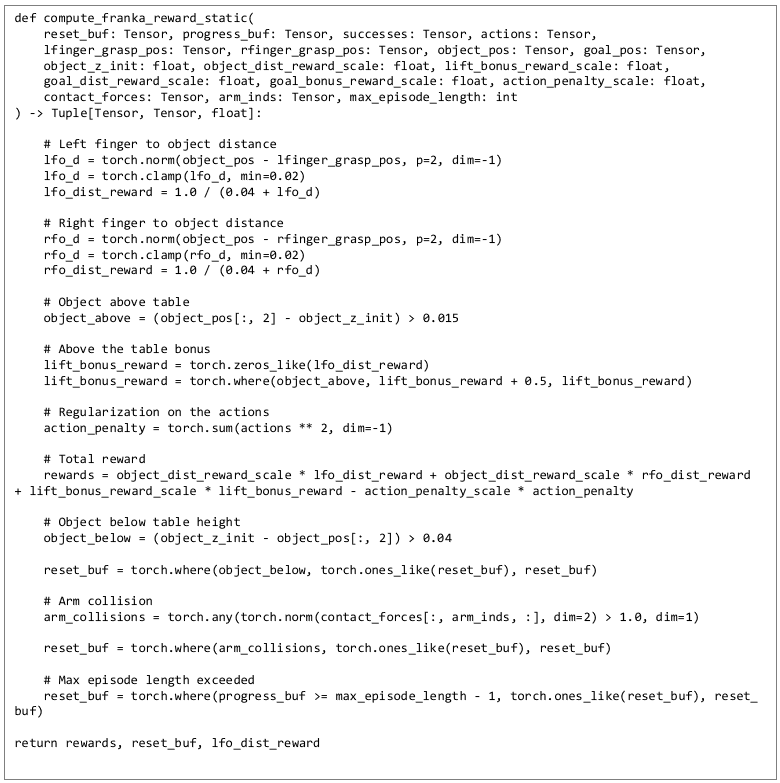}
    \caption{MTRL: Code of the task independent reward component. }
    \label{fig:app.MTRL.85.static}
\end{figure}

\begin{figure}[h]
    \centering
    \includegraphics[width=\linewidth]{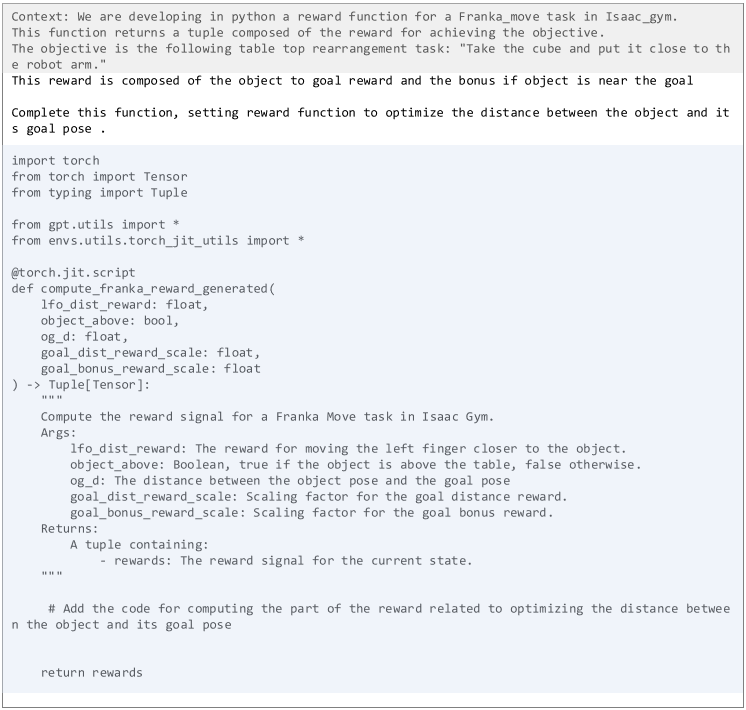}
    \caption{MTRL: Prompt requesting the generation of a task dependent part of a reward function.}
    \label{fig:app.MTRL.86.dynamic.prompt}
\end{figure}

\begin{figure}[ht!]
    \centering
    \includegraphics[width=\linewidth]{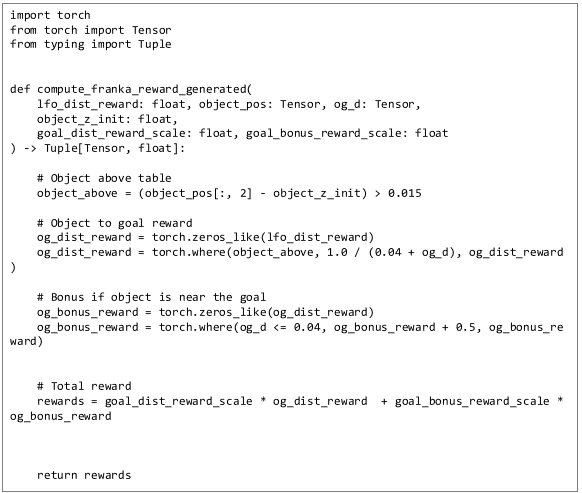}
    \caption{MTRL: Code generated by \larg  \hspace{0.05cm} for the task dependent part of a reward function.}
    \label{fig:app.MTRL.87.dynamic.result}
\end{figure}

The following figures illustrate prompts and results obtained when requesting the generation of ad-hoc code for manipulating one cube to bring it closer to the robotic arm.
Figure ~\ref{fig:app.MTRL.84.global} details the global reward function that combines both elements from the task independent, which is illustrated by figure ~\ref{fig:app.MTRL.85.static}, and task dependent part.  
In this case, \larg focus on generating the dependent part using a prompt illustrated by Figure ~\ref{fig:app.MTRL.86.dynamic.prompt} to produce the code depicted in Figure ~\ref{fig:app.MTRL.87.dynamic.result}.

\clearpage
\newpage

\clearpage
\newpage

\subsubsection{Code validation}
\label{app.code.valid}

\begin{figure}[h]
    \centering
    \includegraphics[width=\linewidth]{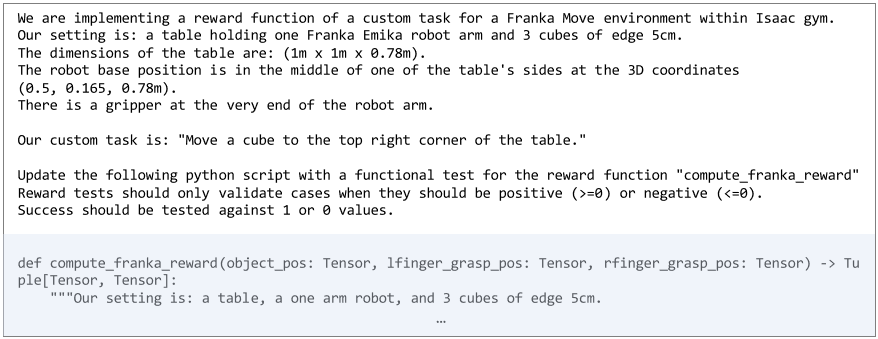}
    \caption{Prompt requesting the generation of functional test for a reward function.}
    \label{fig:app.valid.prompt}
\end{figure}

\begin{figure}[h]
    \centering
    \includegraphics[width=\linewidth]{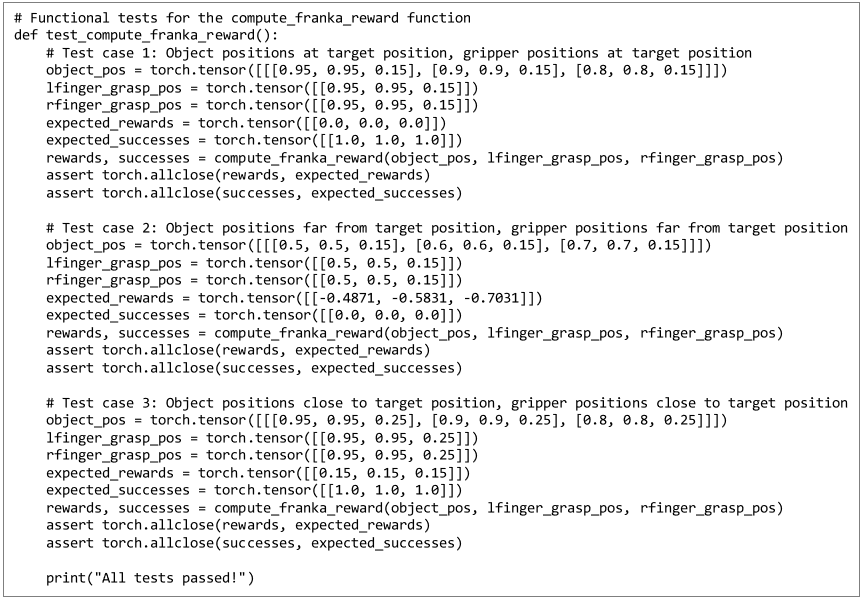}
    \caption{Generated functional test.}
    \label{fig:app.valid.result}
\end{figure}

Optionally, once function code is generated an additional validation step can occur.
LLMs can be used to generate a functional test prior to start the training process or to run the task.
This prompt, illustrated in Figure~\ref{fig:app.valid.prompt}, is composed of 1) a header requesting the LLM to generate a functional test, 2) a list of guidelines to condition the test, and 3) the code of the generated function. 
An example of test is proposed in Figure~\ref{fig:app.valid.result} .

\clearpage
\newpage

\subsection{Experiments}
\label{app.expl}

We evaluate \larg \hspace{0.05cm} on a series of tabletop object  manipulation tasks for both GCRL and MTRL settings.

Experiments leverage the Franka\textunderscore Move environment available on the Isaac\textunderscore Gym repository ~\footnote{https://developer.nvidia.com/isaac-gym}.
This environment describes a table, a Franka Emika Panda robot arm ~\footnote{https://www.franka.de/} which is an open kinematic chain composed with 7DoF, and $n$ cubes on the table.
The dimensions of the table are as follows: 1m x 1m x 0.78m.
The robot arm is placed on the table at (0.5, 0.165, 0.78). 
There is a griper with two fingers attached at the end of the arm.
Cubes with a 5cm edge are located on the surface of the table. 
The global origin (0,0,0) is located on the floor below the table.
Each environment description is written using the Python language.

\subsubsection{Large Language Models}
\label{app.llm}

In our experiments, several LLMs are evaluated: text-davinci-003~\footnote{https://platform.openai.com/docs/models/gpt-3}, code-davinci-002~\footnote{https://platform.openai.com/docs/models/codex} and gpt-3.5-turbo~\footnote{https://platform.openai.com/docs/models/gpt-3-5} from OpenAI which are evolutions from GPT3 optimized with Reinforcement Learning from Human Feedback~\cite{Ouyang2022InstructGPT} .

\begin{figure}[ht!]
    \centering
    \includegraphics[width=\linewidth]{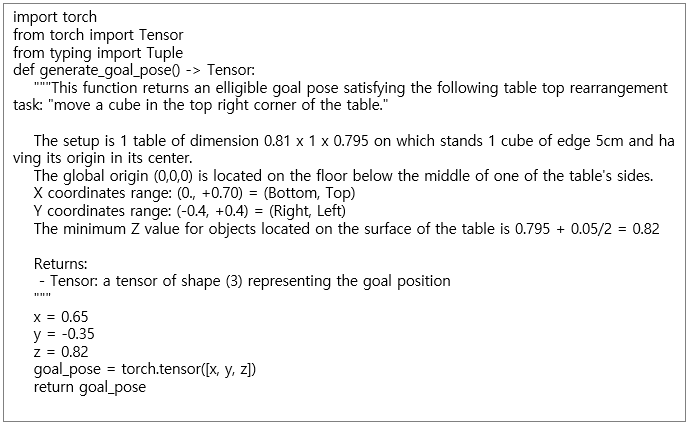}
    \caption{Code generated by gpt-3.5-turbo for the task: Move a cube in the top right corner of the table.}
    \label{fig:app.90}
\end{figure}

\begin{figure}[h!]
    \centering
    \includegraphics[width=\linewidth]{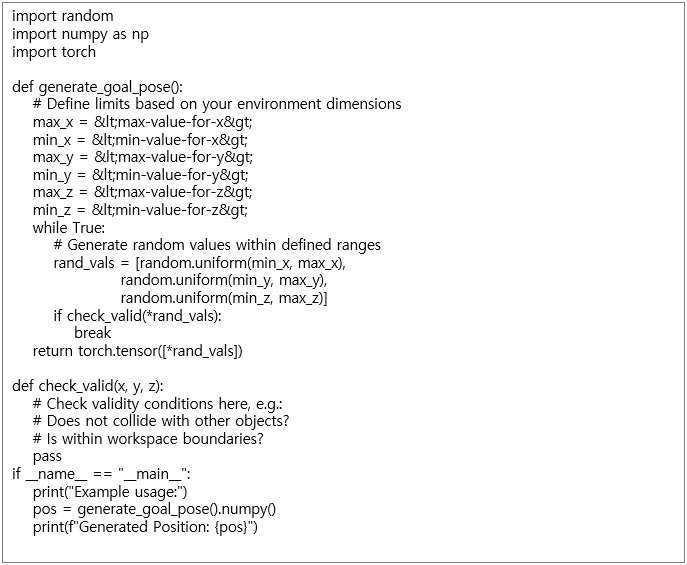}
    \caption{Code generated by StarCoder for the task: Move a cube in the top right corner of the table. In this example the generated code cannot be applied.}
    \label{fig:app.91}
\end{figure}

StarCoder from HuggingFace~\citep{li2023starcoder} is also evaluated to generate goal functions over  the list task defined for the GCRL experiment.
Over 32 tasks, only 12.5\% of generated functions are executable even after the auto-correction loop and only 3.1\% deliver goal positions which are valid.
Frequent issues are related to incorrect variable initialization, missing code and a lack of compliance with provided guidelines such as illustrated in Figures~\ref{fig:app.90} and~\ref{fig:app.91}.

As a summary, best results, in terms of code generation, are achieved with gpt-3.5-turbo which is the model used in results reported  bellow.


\newpage

\subsubsection{Automatic goal generation for the GCRL experiment}
\label{app.gcrl}

In the GCRL experiment, the policy takes as input the position and velocity of each joint of the robot and the respective pose of the objects composing the scene.
The policy trigger joint displacement in a $\mathbb{R}^7$ action space.
In addition to the position of the object composing the scene, the policy takes as input the goal positions. These positions are provided by goal functions generated by \larg.
The policy is trained beforehand using Proximal Policy Optimization ~\cite{Schulman2017ProximalPO} with default Franka\_Move parameters as defined in table ~\ref{tab:ppo_param.list}.

\begin{table*}[ht!]
\small
\centering
\begin{tabular}{|l|l|}
\hline
training parameters 	& values \\
\hline
number of environments 	& 2048 \\
episode length 	& 500 \\
object distance reward scale 	& 0.08 \\
lift bonus reward scale 	& 4.0 \\
goal distance reward scale 	& 1.28 \\
goal bonus reward scale 	& 4.0 \\
action penalty scale 	& 0.01 \\
collision penalty scale 	& 1.28 \\
actor hidden dimension 	& [256, 128, 64] \\
critic hidden dimension 	& [256, 128, 64] \\
\hline
\end{tabular}
\caption{List of parameters used in the Franka\_Move PPO training loop.}
\label{tab:ppo_param.list}
\end{table*}

\begin{figure}[ht!]
    \centering
    \includegraphics[width=\linewidth]{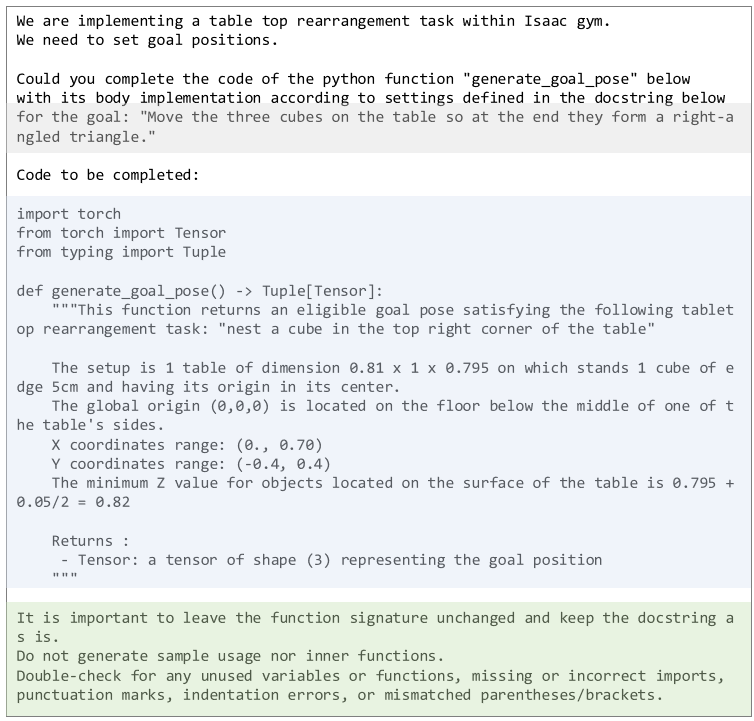}
    \includegraphics[width=\linewidth]{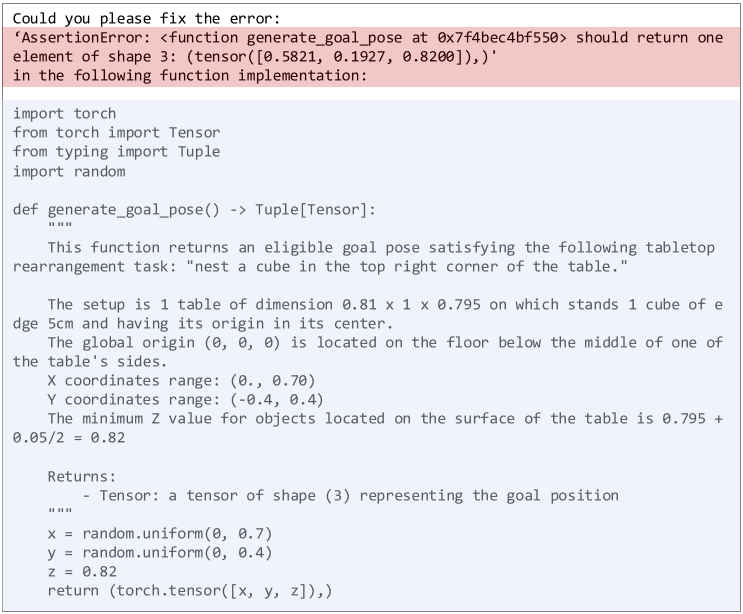}
    \includegraphics[width=\linewidth]{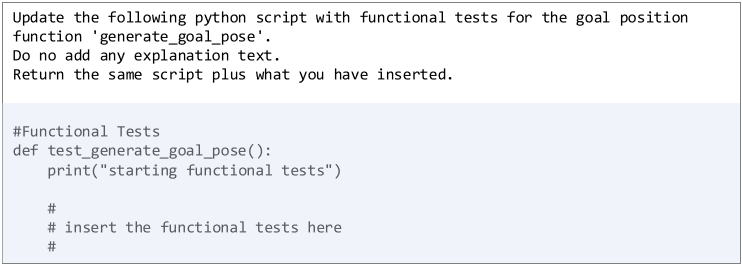}
    \caption{Prompts illustrating the 3 steps involved in the generation of a valid goal positioning function: 1) request to generate a function according to specific environment parameters, 2) auto-correction, 3) final validation. The highlighted section in red contains the error message generated at the execution phase.}
    \label{fig:60}
\end{figure}

We evaluate our approach on a series of 32 tasks including 27 tasks involving a single object, and 5 tasks involving 3 objects. 
Tasks $d17$ to $d27$ correspond to objectives which are defined relative to the object's initial position.
In this case, the signature of the  goal function naturally takes as input the initial position of the cubes composing the scene.
Figure ~\ref{fig:60} illustrates the prompting workflow which translates a task description into the generation of a goal  function. 
It involves an auto-correction step and the production of a functional test afterward.

\begin{figure}[ht!]
    \centering
    \includegraphics[width=\linewidth]{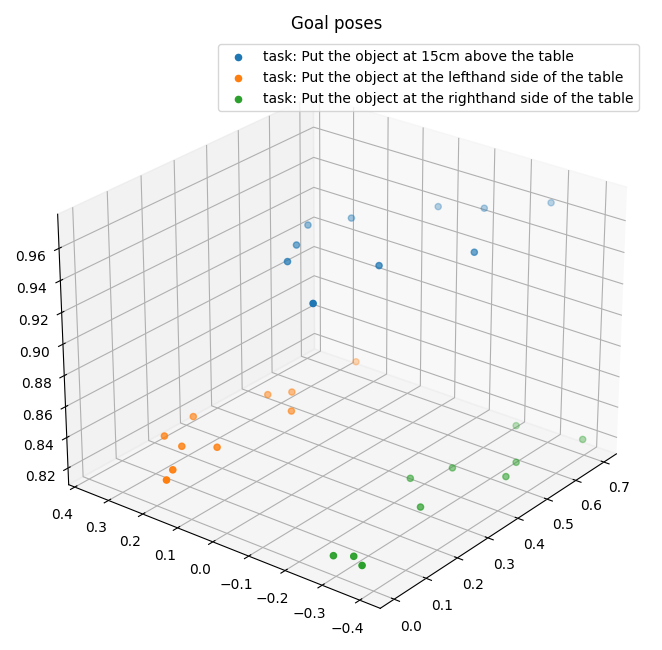}
    \caption{Example of goal positions generated by our method for 3 different tasks requesting targets to be located on the right, left, and above the table.}
    \label{fig:61}
\end{figure}

Figure~\ref{fig:61} illustrates results produced by 10 run of 3 different goal functions generated out of 3 different manipulation tasks
In all cases, the resulting poses are well aligned with task requirements while exploring the range of valid positions allowed by a non deterministic task definition.

Table~\ref{tab:1cube.listtask} provides the list of all tasks used in our experiment and report the compliance of generated goals with task descriptions.

\begin{table*}[ht!]
\small
\centering
\begin{tabular}{|l|p{8cm}|c|}
\hline

ID & Task & Generated Pose validity   \\
\hline
d01 & Move a cube to the top right corner of the table. & \checkmark  \\
d02 & Move a cube to the top left corner of the table. & \checkmark  \\
d03 & Move a cube to the bottom right corner of the table. & \checkmark  \\
d04 & Move a cube to the bottom left corner of the table. & \checkmark  \\
d05 & Lift the cube 15cm above the table. & \checkmark  \\
d06 & Rotate a cube upside-down. & \checkmark  \\
d07 & Take to cube and move it to the left side of the table. & -  \\
d08 & Take to cube and move it to the right edge of the table. & \checkmark  \\
d09 & Take to cube and raise it at 20 cm to the far side of the table.  & \checkmark  \\
d10 & Take the cube and move it closer to the robotic arm. & \checkmark  \\
d11 & Pick up the cube and move it away from the robotic arm. & \checkmark  \\
d12 & Take the cube and move it very close to the robotic arm. & -  \\
d13 & Push the cube off the limits of the table. & \checkmark  \\
d14 & Bring the cube closer to the robot arm. & \checkmark  \\
d15 & Move the cube to one corner of the table. & \checkmark  \\
d16 & Place the cube anywhere on the diagonal of the table running from the top right corner to the bottom left corner. & \checkmark  \\
\hline
d17 & Lift the cube 15cm above the table and 10 cm to the right. & \checkmark  \\
d18 & Lift the cube 20cm above the table and 15 cm ahead. & \checkmark  \\
d19 & Lift the cube 20cm above the table and 15 cm backward. & \checkmark  \\
d20 & Push a cube 10cm to the right and 10cm ahead. & \checkmark  \\
d21 & Push a cube 10cm to the right and 10cm backward. & \checkmark  \\
d22 & Push a cube 10cm to the left and 10cm ahead. & \checkmark  \\
d23 & Push a cube 10cm to the left and 10cm backward & \checkmark  \\
d24 & Grab a cube and move it a bit to the left. & \checkmark  \\
d25 & Grab a cube and lift it a bit and move it a bit ahead. & \checkmark  \\
d26 & Move the cube at 20cm to the left of its initial position. & \checkmark  \\
d27 & Move the cube 20cm above its current position. & \checkmark  \\
\hline
d28 & Move one cube to the left side of the table, another one to the right side of the table, and put the last cube at the center of the table. & \checkmark \\
d29 & Move the three cubes so they are 10 cm close to one another. & \checkmark  \\
d30 & Move the three cubes on the table so that at the end they form a right-angled triangle. & \checkmark  \\
d31 & Move the three cubes on the table so that at the end they form an isosceles triangle. & \checkmark  \\
d32 & Reposition the three cubes on the table such that they create a square, with the table's center serving as one of the square's corners. & \checkmark  \\
\hline
\end{tabular}
\caption{List of the 32 manipulation tasks evaluated with \larg. Tasks $d17$ to $d27$ involve objectives relative to the object's initial position.
Tasks $d28$ to $d32$ address 3 object manipulation problems and therefore 3 goals. Localisation compliance with task definition is  reported.}
\label{tab:1cube.listtask}
\end{table*}

\begin{figure}[!htb]
     \centering
     \includegraphics[width=.5\linewidth]{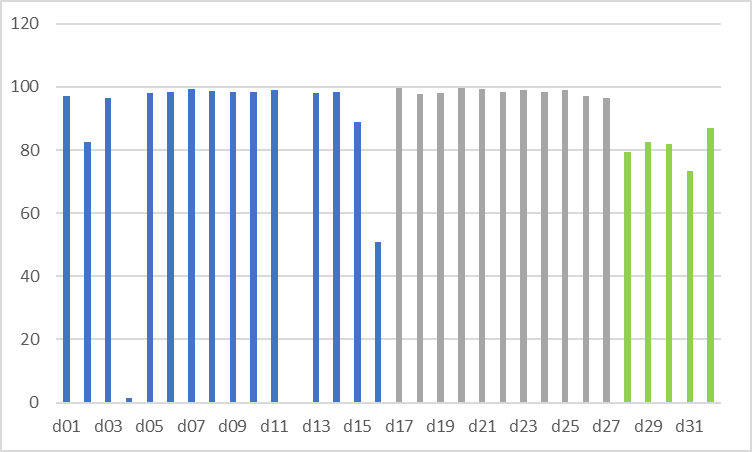}
     \caption{Success rate for GCRL manipulation tasks. Blue reflects 1 object manipulation for absolute pose whereas grey reflects relative object pose. Green relates to 3 object manipulation tasks.}
     \label{fig:app.70}
\end{figure}

Figure~\ref{fig:app.70} shows  success rates for our 32 manipulation tasks.
Looking at unsuccessful experiments, we make several  observations. 
A common source of error relates to a lack of contextual information and constraints in the definition of the task.
Two options can be mentioned as future directions to address such a case: either to increase the amount of constraints in the prompt, or to use a  LLM with more capabilities. 

\begin{figure}[ht!]
    \centering
    \includegraphics[width=\linewidth]{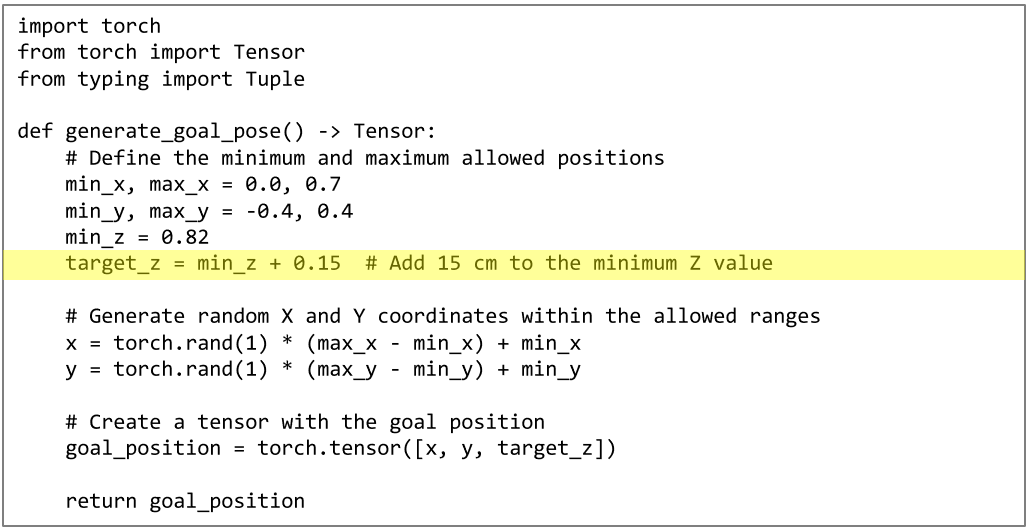}
    \caption{Arithmetic capabilities of the LLM for Task \textit{d05}. The comment highlighted in yellow so as the related code is generated by the LLM. }
    \label{fig:app.71}
\end{figure}

Interestingly, this experiment also allows to highlight LLM reasoning capabilities as illustrated in Figure~\ref{fig:app.71} where the task request to lift a cube at 15cm above the table. 
In this case the generated goal function add the table height to the specified 15cm to end up with the correct position.

As a summary, \larg  \hspace{0.05cm} allows to generate code for goal prediction according to textual task descriptions. 
In some cases the generated code do not properly fit with user specifications but our experiment demonstrate that a feedback loop with additional guidelines can fix the problem.

\clearpage
\newpage

\subsubsection{Automatic reward generation for the MTRL experiment}

Our second experiment evaluates \larg \hspace{0.05cm} capability to address MTRL settings.
For task encoding, we use the Google T5-small language model.
We use the [CLS] token embedding computed by the encoder stack of the model which is defined in $\mathbb{R}^{512}$.
We concatenate this embedding with the state information of our manipulation environment defined in $\mathbb{R}^{7}$ and feed it into a fully connected network stack used as policy.
This policy is composed of three layers using respectively, $\{512, 128, 64\}$ hidden dimensions.

\begin{figure}[ht!]
   \begin{minipage}{0.48\textwidth}
     \centering
    \includegraphics[width=\linewidth]{Fig.81.MTRL.success.rate.png}
    \caption{Success rate evaluations of MTRL over automatic reward generation.}
    \label{fig:app.82}
   \end{minipage}\hfill
   \begin{minipage}{0.48\textwidth}
     \centering
     \includegraphics[width=.8\linewidth]{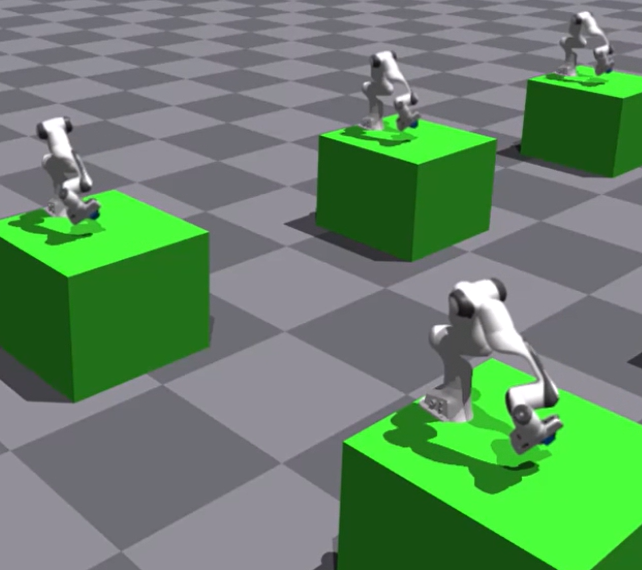}
    \caption{Example of multi-lingual capabilities for robot manipulation. In our simulation tasks are submitted using different languages including English, Arabic and Korean. This figure illustrate task $m04$ translated in Korean.}
    \label{fig:app.82.2}
   \end{minipage}

\end{figure}

In our experiment we train an MTRL settings using  Proximal Policy Optimization  with default Franka Move parameters using reward functions generated by \larg over 9 tasks listed in Table~\ref{tab:app.1cube.MTRL.listtask}.
These tasks address one object manipulation on a tabletop. 
We leverage the LLM capabilities to paraphrase these tasks to produce the evaluation set.
Paraphrases include task translation as the T5 model is trained for down stream tasks such as machine translation. Figure~\ref{fig:app.82.2} illustrates the application of Task $m04$ submitted as a text based command in Korean language \begin{CJK}{UTF8}{mj} ("큐브를 테이블 중앙으로부터 20cm 위로 옮겨주세요") \end{CJK} to a policy trained in MTRL.


\begin{table*}[ht!]
\small
\centering
\begin{tabular}{|l|m{8cm}|}
\hline

ID & Task   \\
\hline
m01 & Push the cube to the far right of the table.  \\
m02 & Move a cube to the top left corner of the table.  \\
m03 & Take the cube and put it close to the robot arm.  \\
m04 & Move a cube at 20cm above the center of the table.  \\
m05 & Move a cube at 15 cm above the table.  \\
m06 & Take the cube and put it on the diagonal of the table.  \\
m07 & Push the cube at 20cm ahead of its current position.  \\
m08 & Move the cube to the center of the table.  \\
m09 & Grab the cube and move it forward to the left.   \\
\hline
\end{tabular}
\caption{List of task used in the MTRL settings. }
\label{tab:app.1cube.MTRL.listtask}
\end{table*}

Figure ~\ref{fig:app.82} provides success rates obtained for the 9 tasks used in the MTRL experiment.
It illustrates \larg \hspace{0.05cm} capability to generate valid reward functions  to train and execute MTRL policies conditioned by textual task definitions.


\clearpage
\newpage

\subsection{Examples of goal functions generated by \larg}


\begin{figure}[h]
    \centering
    \includegraphics[width=\linewidth]{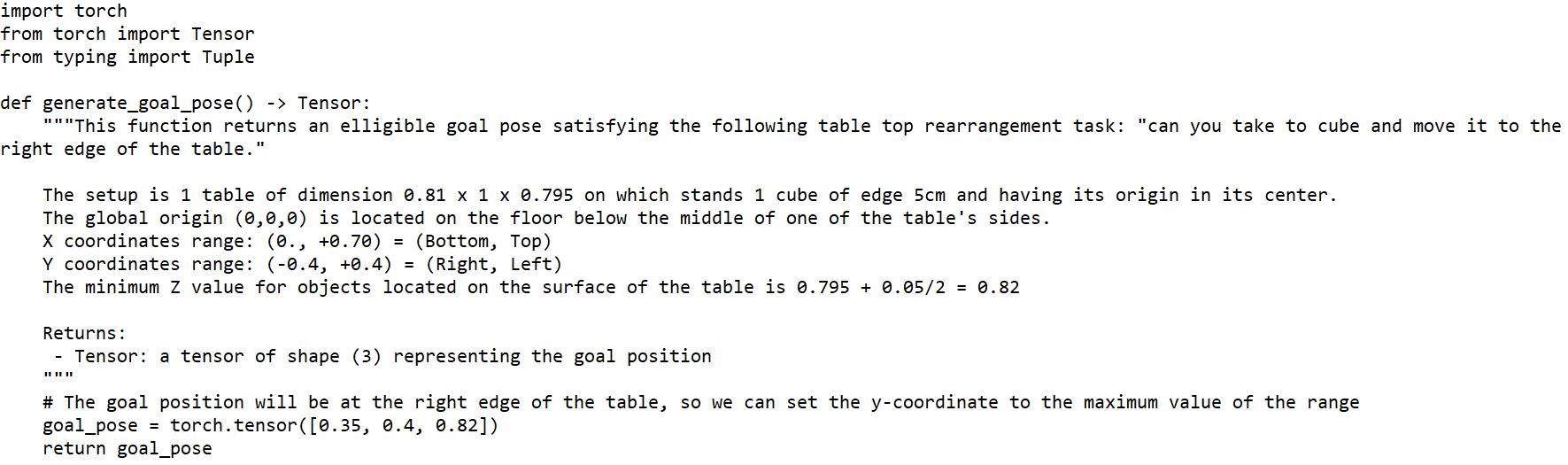}
    \caption{Task $d08$: Take to cube and move it to the right edge of the table}
    \label{fig:app:T8}
\end{figure}

\begin{figure}[h]
    \centering
    \includegraphics[width=\linewidth]{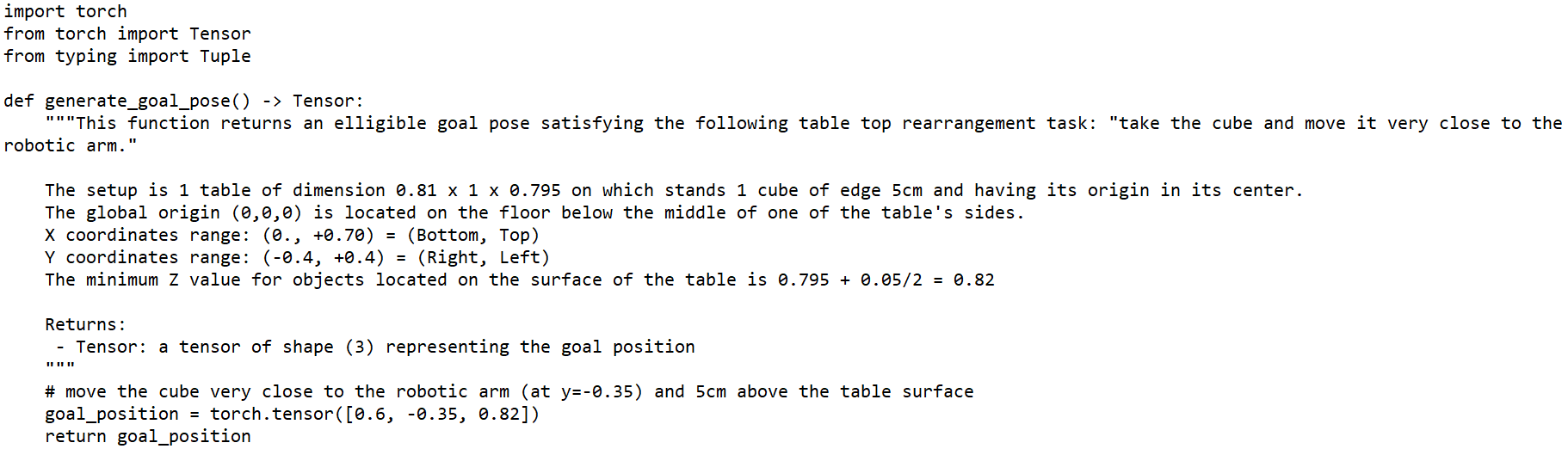}
    \caption{Task $d12$: Take the cube and move it very close to the robotic arm.}
    \label{fig:app:T12}
\end{figure}

\begin{figure}[h]
    \centering
    \includegraphics[width=\linewidth]{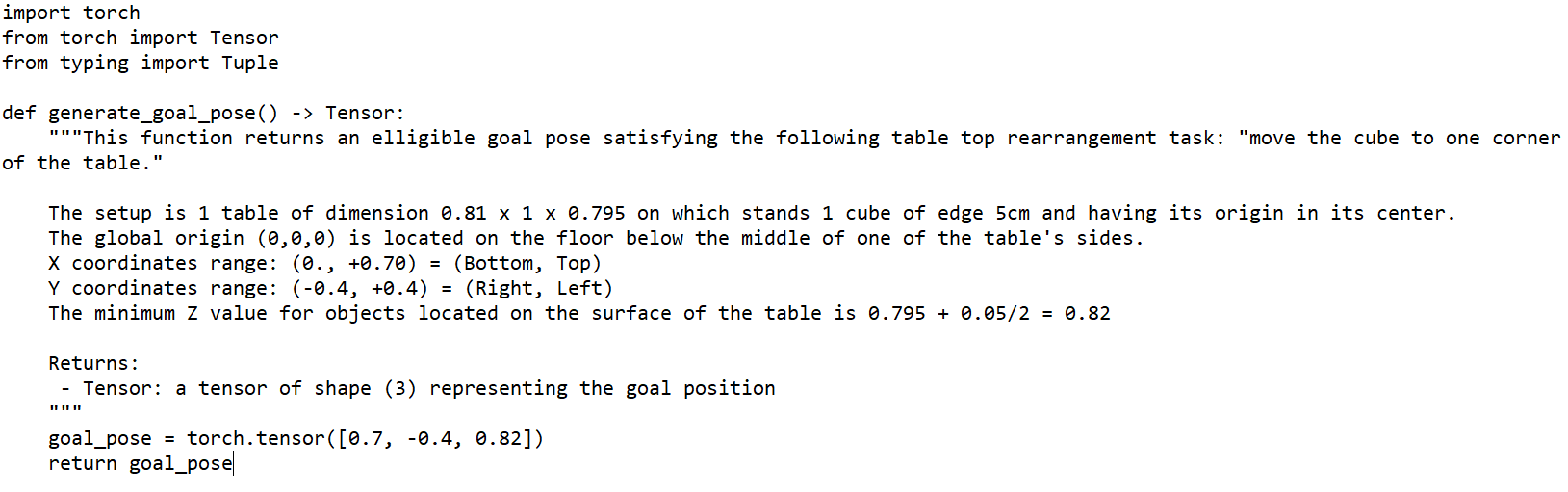}
    \caption{Task $d15$: Move the cube to one corner of the table.}
    \label{fig:app:T15}
\end{figure}

\begin{figure}[h]
    \centering
    \includegraphics[width=\linewidth]{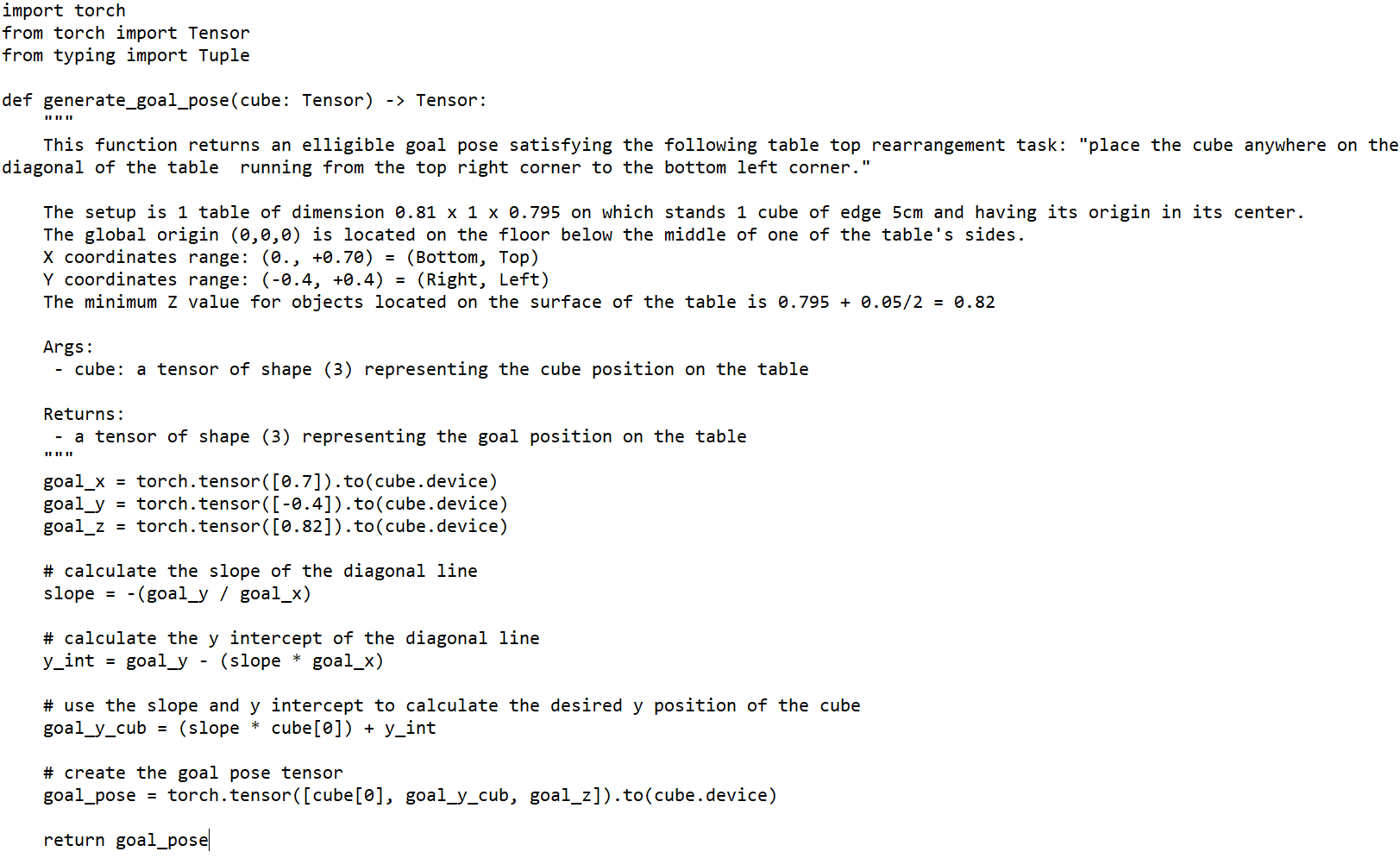}
    \caption{Task $d16$: Place the cube anywhere on the diagonal of the table  running from the top right corner to the bottom left corner.}
    \label{fig:app:T16}
\end{figure}

\begin{figure}[h]
    \centering
    \includegraphics[width=\linewidth]{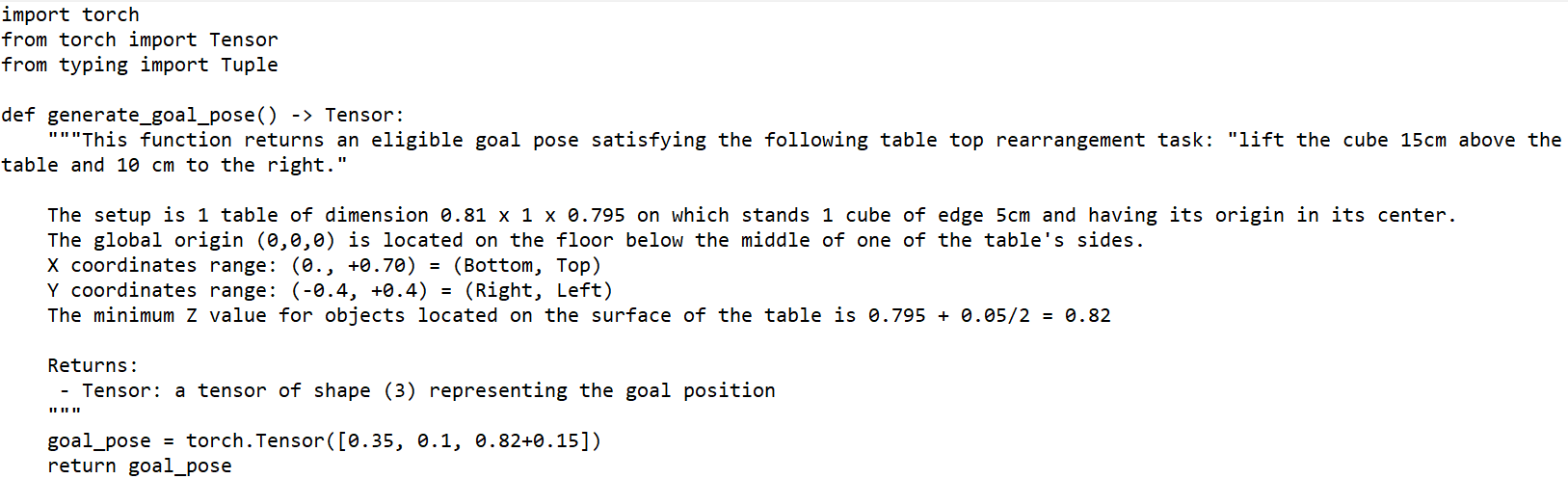}
    \caption{Task $d17$: Lift the cube 15cm above the table and 10 cm to the right.}
    \label{fig:app:T17}
\end{figure}

\begin{figure}[h]
    \centering
    \includegraphics[width=\linewidth]{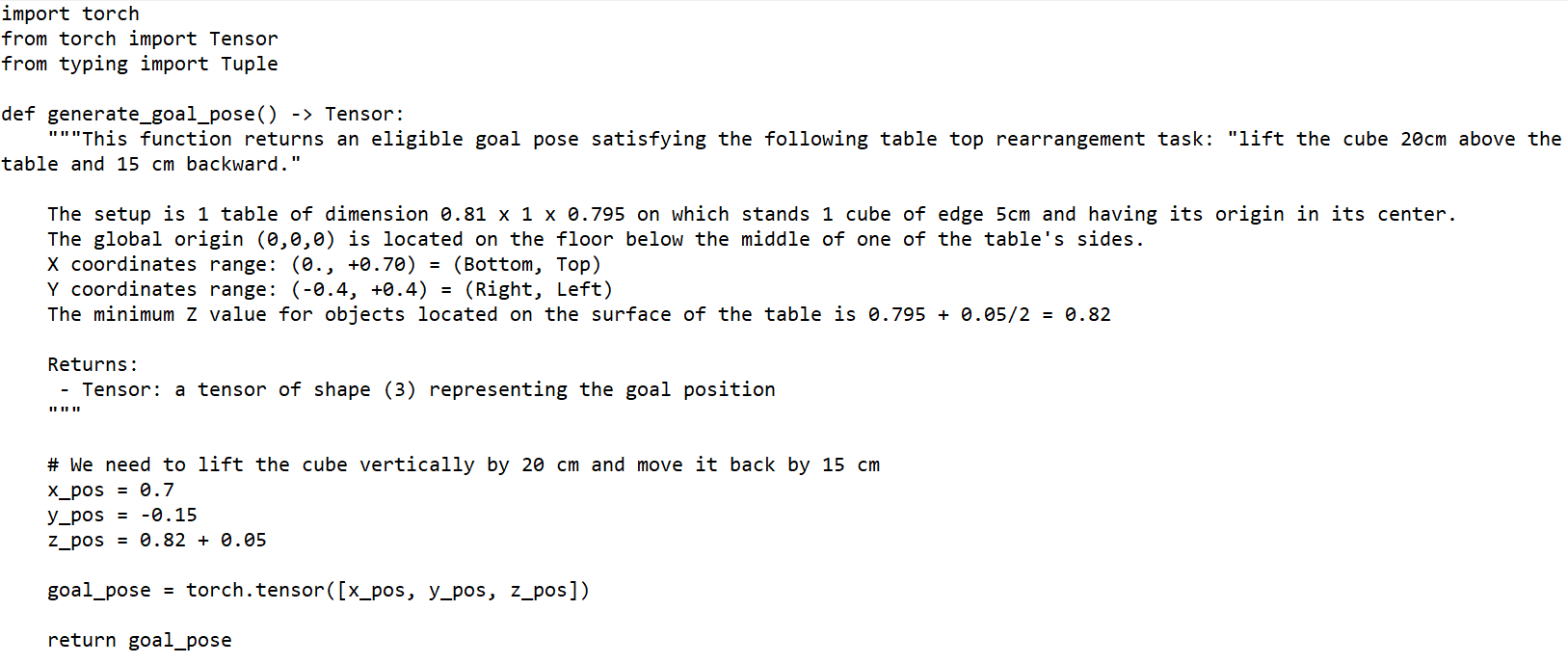}
    \caption{Task $d19$: Lift the cube 20cm above the table and 15 cm backward.}
    \label{fig:app:T19}
\end{figure}

\begin{figure}[h]
    \centering
    \includegraphics[width=\linewidth]{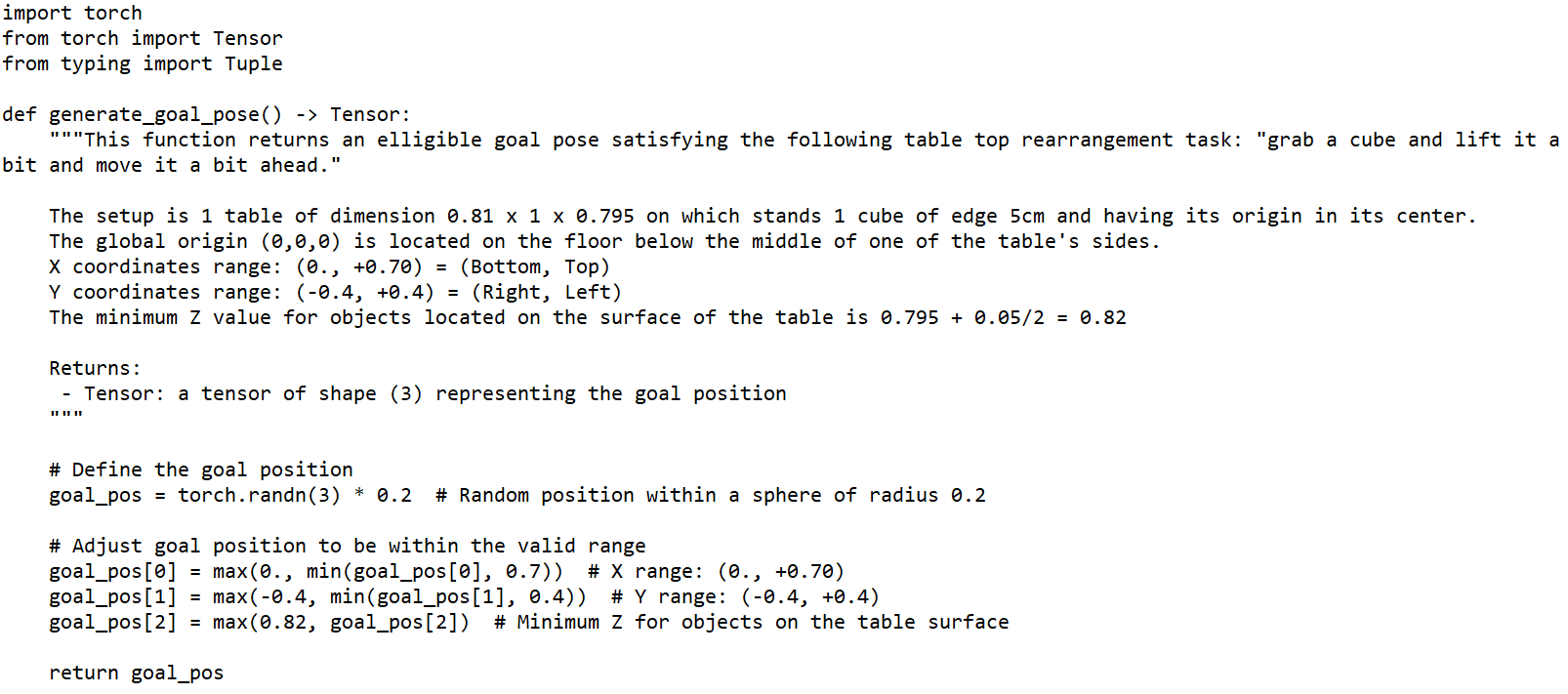}
    \caption{Task $d25$: Grab a cube and lift it a bit and move it a bit ahead.}
    \label{fig:app:T25}
\end{figure}

\begin{figure}[h]
    \centering
    \includegraphics[width=\linewidth]{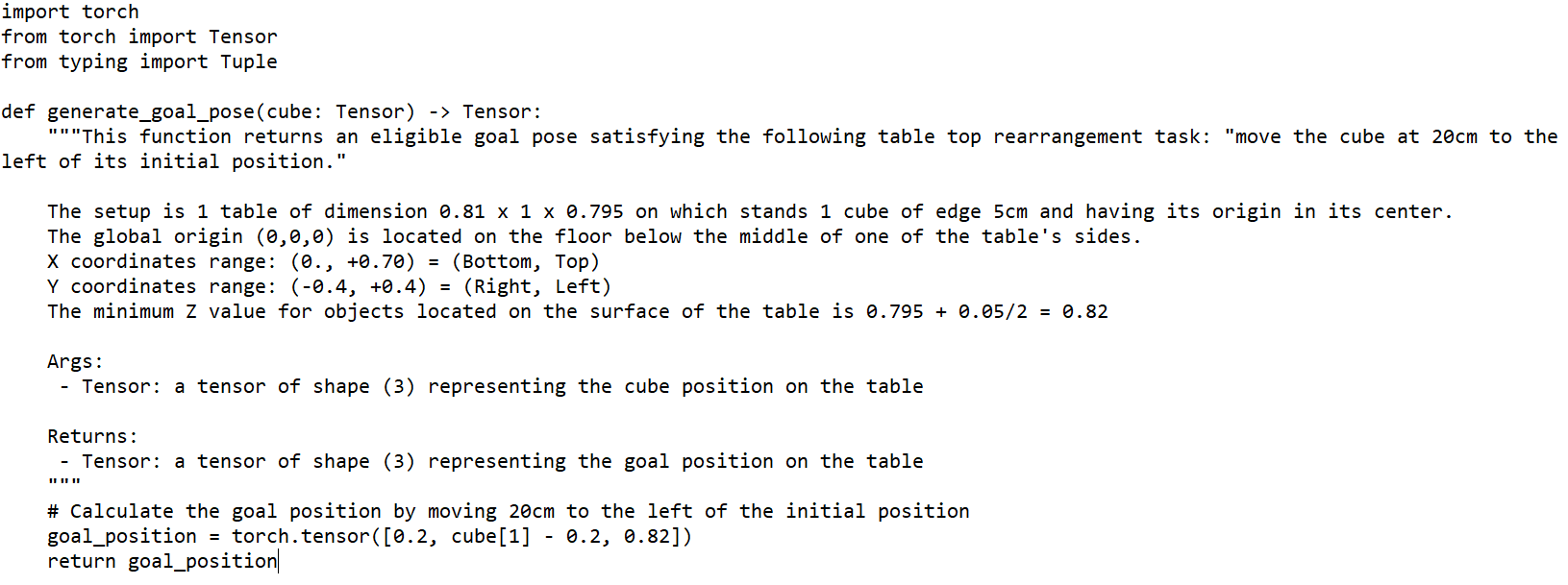}
    \caption{Task $d26$: Move the cube at 20cm to the left of its initial position.}
    \label{fig:app:T26}
\end{figure}

\begin{figure}[h]
    \centering
    \includegraphics[width=\linewidth]{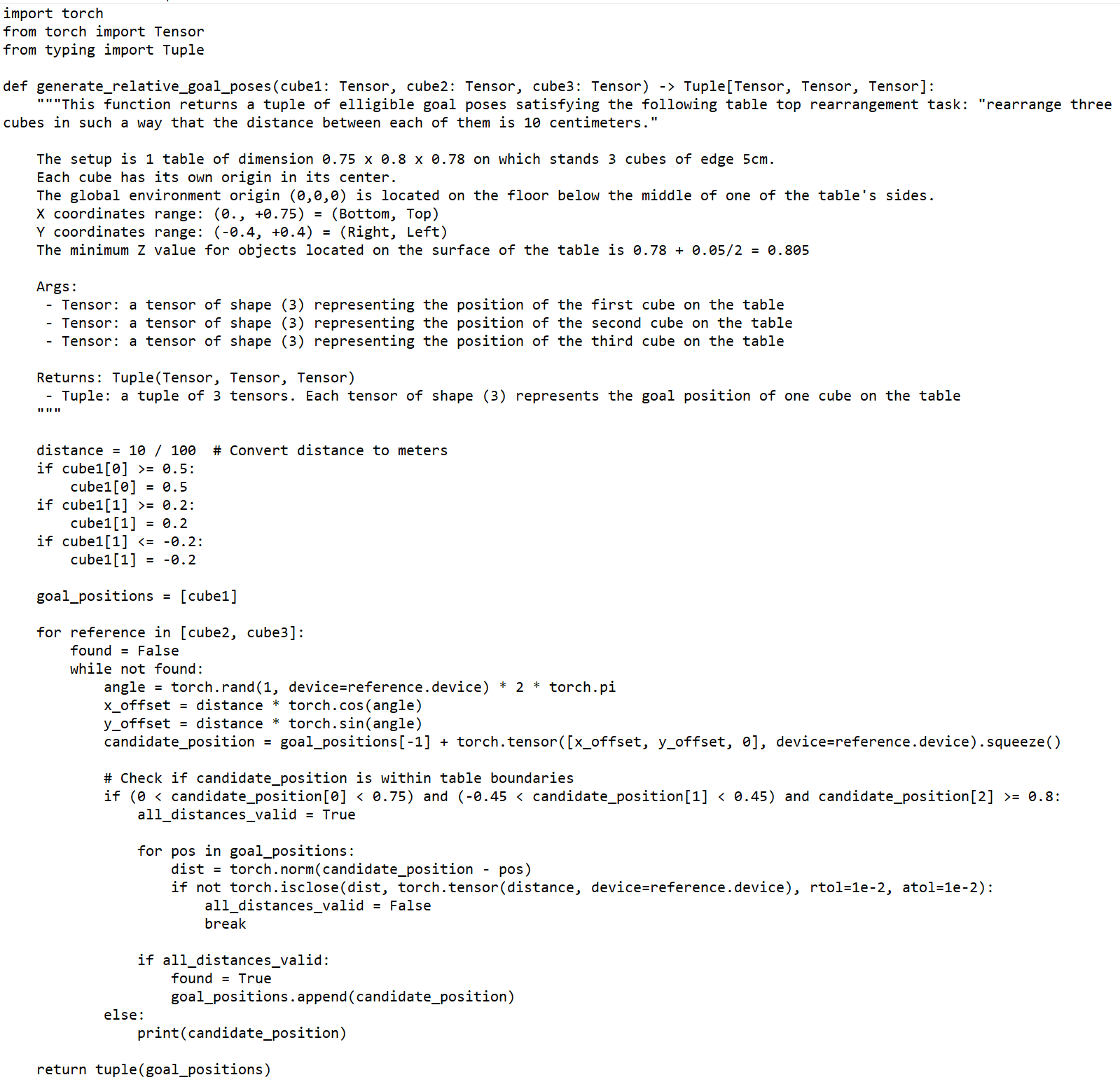}
    \caption{Task $29$: Rearrange three cubes in such a way that the distance between each of them is 10 centimeters.}
    \label{fig:app:T29}
\end{figure}

\begin{figure}[h]
    \centering
    \includegraphics[width=\linewidth]{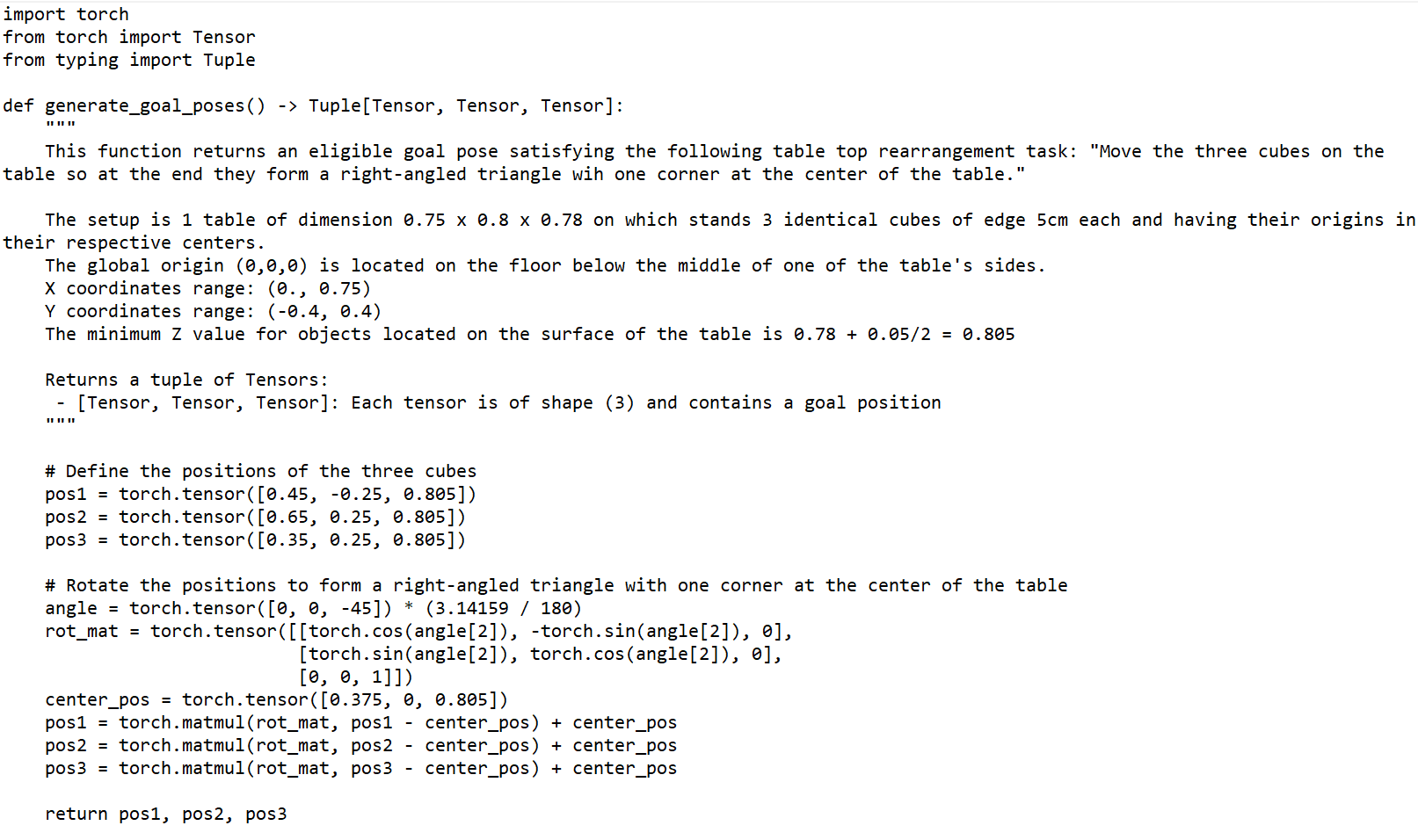}
    \caption{Task $d30$: Move the three cubes on the table so at the end they form a right-angled triangle with one corner at the center of the table.}
    \label{fig:app:T30}
\end{figure}

\end{document}